\documentclass[10pt,twocolumn,letterpaper]{article}

\usepackage[pagenumbers]{cvpr} %

\definecolor{cvprblue}{rgb}{0.21,0.49,0.74}
\definecolor{myorange}{RGB}{255,128,0}
\usepackage[pagebackref,breaklinks,colorlinks,allcolors=cvprblue]{hyperref}
\usepackage[normalem]{ulem}
\usepackage{bxcoloremoji}
\title{\tracegenlogo TraceGen: World Modeling in 3D Trace-Space Enables Learning from Cross-Embodiment Videos}

\author{
Seungjae Lee$^{1}$\thanks{Equal contribution} \quad
Yoonkyo Jung$^{1}$\footnotemark[1] \quad
Inkook Chun$^{2}$\footnotemark[1] \quad
Yao-Chih Lee$^{1}$ \quad
Zikui Cai$^{1}$ \\
Hongjia Huang$^{2}$ \quad
Aayush Talreja$^{1}$ \quad
Tan Dat Dao$^{1}$ \quad
Yongyuan Liang$^{1}$ \\
Jia-Bin Huang$^{1}$ \quad
Furong Huang$^{1}$ \\ \\
$^{1}$University of Maryland, College Park \quad
$^{2}$New York University \\ \\
\vspace{2mm}
Project Page: \href{https://tracegen.github.io}{\textcolor{myorange}{https://tracegen.github.io}}
}

\usepackage{xcolor}
\definecolor{darkgreen}{RGB}{0,120,0} %
\usepackage{tcolorbox}
\tcbuselibrary{skins, breakable}
\usepackage{array} %
\usepackage{booktabs}
\usepackage{cuted}
\usepackage{multirow}
\usepackage{graphicx}

\newcommand{\tracegenlogo}{%
    \raisebox{-0.2em}{\includegraphics[height=1.2em]{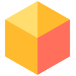}}%
}

\newcommand{\traceforgelogo}{%
    \!%
    \raisebox{-0.2em}{\includegraphics[height=1.2em]{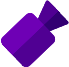}}%
    \hspace{-0.3em}
}

\begin{document}
\maketitle

\begin{strip}
\centering
\vspace{-3.5em}
\includegraphics[width=1.\linewidth]{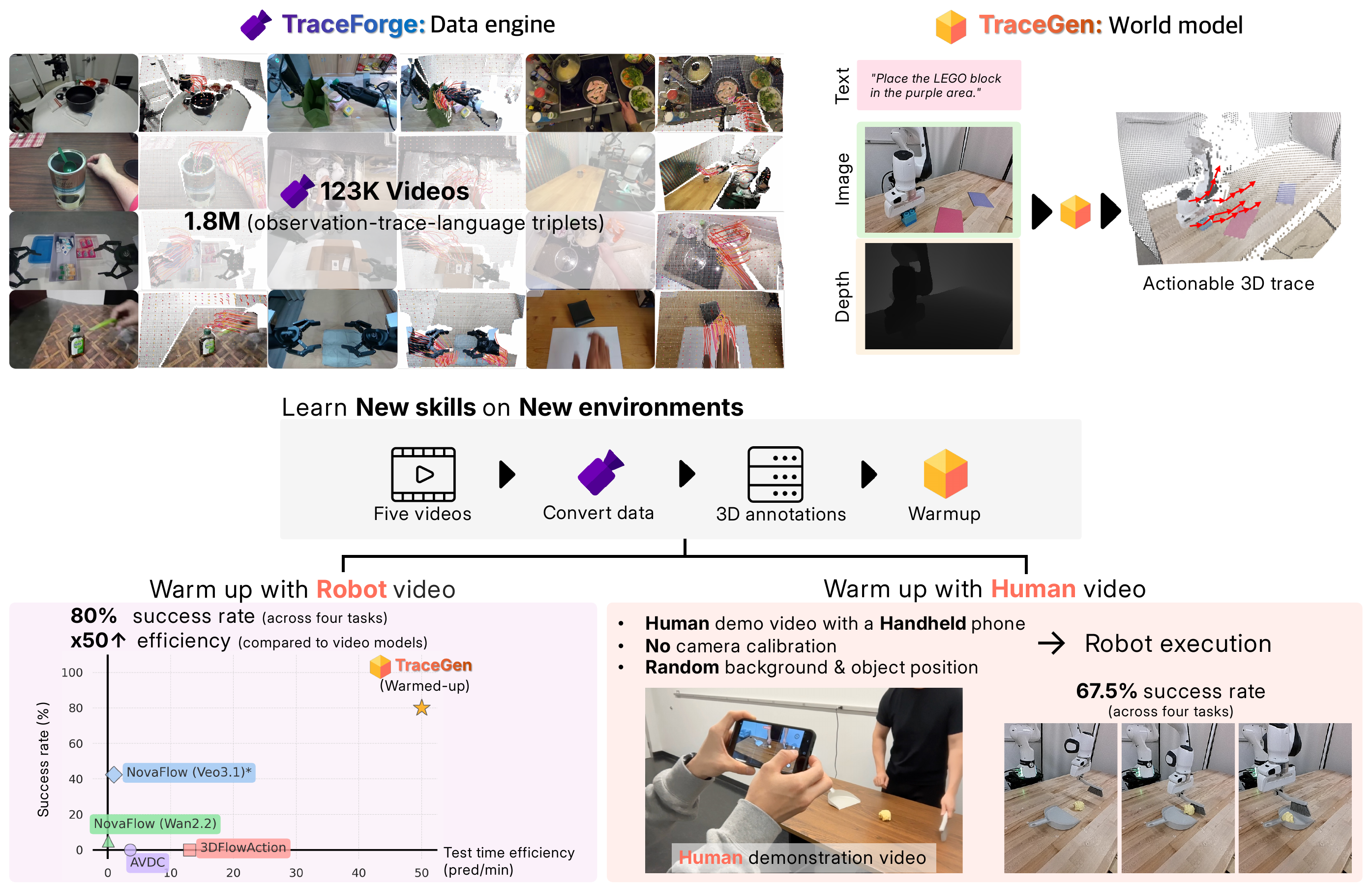}
\captionof{figure}{
\traceforgelogo \textbf{TraceForge} provides the structured training signal, and \tracegenlogo \textbf{TraceGen} consumes this signal to learn a world model in 3D trace space. 
Pretrained on 1.8M observation–trace–language triplets from the TraceForge-123K corpus---combining in-the-wild human videos and heterogeneous robot datasets---TraceGen acquires a strong 3D motion prior, enabling rapid adaptation to new skills and new environments.
\emph{Bottom-left}: \textbf{Robot-domain warm-up}. With only five target-robot demonstrations, TraceGen reaches \textbf{80\% success} across four tasks and is \textbf{50$\times$ faster} than video-based world models (Veo~3.1 inference via API averages).
\emph{Bottom-right}: \textbf{Human$\rightarrow$Robot transfer}. With just five uncalibrated handheld human videos---featuring different backgrounds and object positions---TraceGen attains \textbf{67.5\%} real-robot success.
}
\label{fig:thumbnail}
\end{strip}

\begin{abstract}

Learning new robot tasks on new platforms and in new scenes from only a handful of demonstrations remains challenging. While videos of \emph{other embodiments}---humans and different robots---are abundant, differences in embodiment, camera, and environment hinder their direct use. We address the \emph{small-data} problem by introducing a unifying, symbolic representation---a compact 3D ``trace-space'' of scene-level trajectories---that enables learning from cross-embodiment, cross-environment, and cross-task videos. We present \tracegenlogo \textbf{TraceGen}, a world model that predicts future motion in trace-space rather than pixel space, abstracting away appearance while retaining the geometric structure needed for manipulation. To train TraceGen at scale, we develop \traceforgelogo \textbf{TraceForge}, a data pipeline that transforms heterogeneous human and robot videos into consistent 3D traces, yielding a corpus of 123K videos and 1.8M observation--trace--language triplets. Pretraining on this corpus produces a transferable 3D motion prior that adapts efficiently: with just five target robot videos, TraceGen attains $80\%$ success across four tasks while offering $50$--$600\times$ faster inference than state-of-the-art video-based world models. In the more challenging case where only five uncalibrated human demonstration videos captured on a handheld phone are available, it still reaches $67.5\%$ success on a real robot, highlighting TraceGen’s ability to adapt across embodiments without relying on object detectors or heavy pixel-space generation.
\end{abstract}

\section{Introduction}
\label{sec:intro}

Robots are expected to master diverse manipulation tasks across platforms and scenes, yet collecting sufficient, task-specific robot demonstrations is slow and costly. In contrast, large corpora of human videos are readily available, but embodiment, camera, and scene disparities make direct reuse difficult. We ask: \emph{Can we exploit cross-embodiment videos to overcome small-data regimes for new robots and tasks?}

\paragraph{Limitations of pixel and language spaces.}
Recent progress in large vision-language-action models and multitask policies is notable, but performance often degrades outside training domains \cite{kim2024openvla, bjorck2025gr00t, barreiros2025careful}. A natural alternative is to leverage pretrained world models built on video generation or vision-language models (VLMs) \cite{li2025hamster, li2025novaflow, yuan2025embodied, qu2025embodiedonevision, yuan2024robopoint, yuan2025seeing}. However, video generators operate in \emph{pixel space}, allocating capacity to backgrounds and textures that are irrelevant to control, while VLMs produce token sequences that lack the spatial precision required for fine-grained object motion. In both families, inference is computationally expensive, complicating real-time planning and fine-tuning.

\paragraph{Key insight: a shared 3D structure.}
Although embodiments differ in kinematics and scale, the motion of manipulated objects and end-effectors admits a shared, scene-centric 3D structure. We term this compact, symbolic representation the \emph{trace-space}: a sequence of 3D trajectories that captures the \emph{where and how} of motion while discarding appearance and backgrounds. Learning in trace-space promises invariance to camera and environment, and a practical path to reusing cross-embodiment, in-the-wild video.

\paragraph{Approach: TraceGen in trace-space.}
We propose \tracegenlogo \textbf{TraceGen}, a world model that predicts future motion directly in 3D trace-space rather than pixels. By modeling scene-level trajectories, TraceGen focuses on the geometric signal pertinent to manipulation and avoids heavy generative rendering. 
Pretraining on in-the-wild human videos and heterogeneous robot datasets gives TraceGen a transferable motion prior that adapts to new robots and scenes with only a few warm-up videos—enabling fast human$\rightarrow$robot and robot$\rightarrow$robot transfer without object detectors or heuristic filtering.
\begin{figure}[t!]
  \centering
  \includegraphics[width=1.00\linewidth]{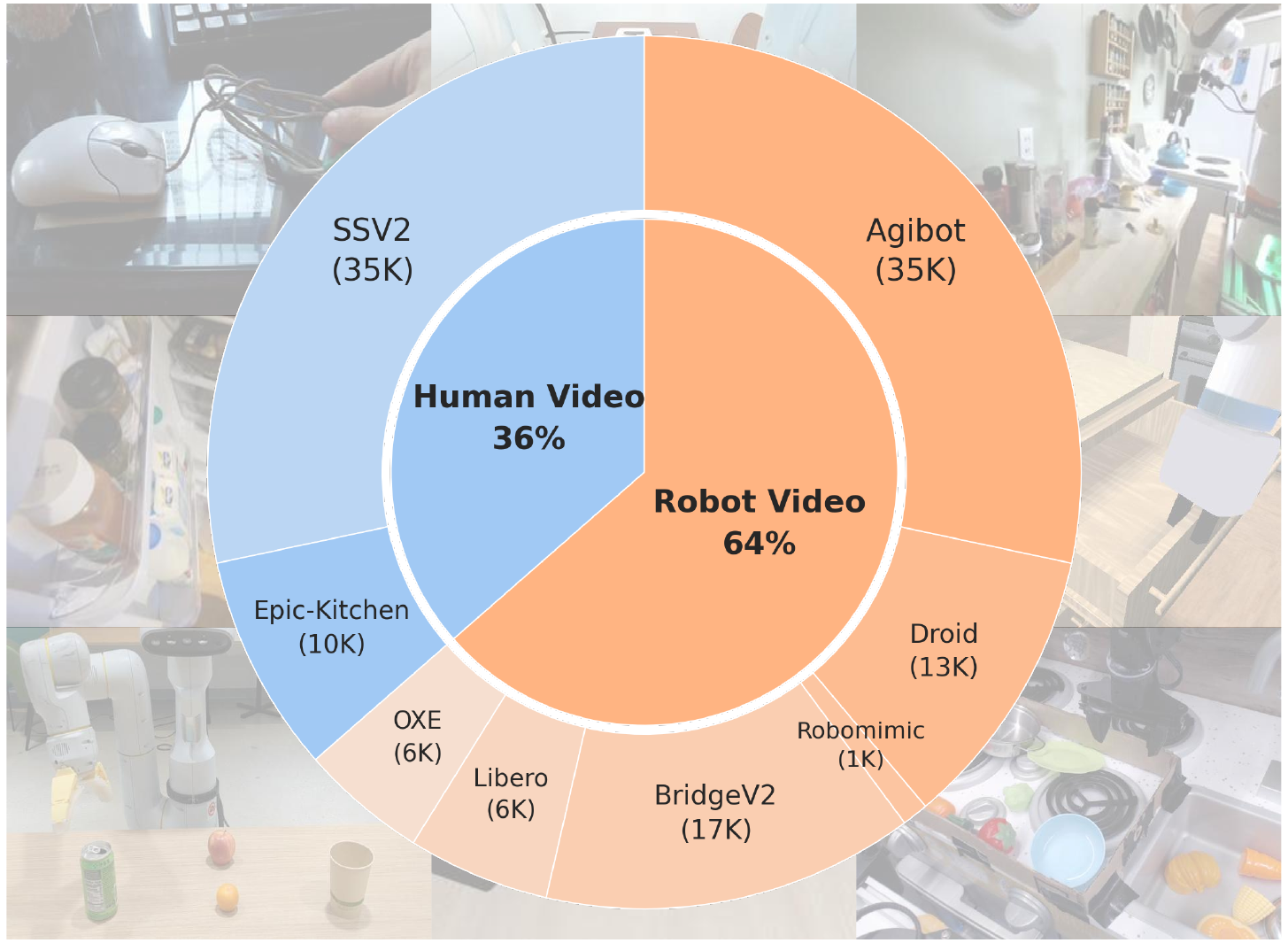}
  \caption{\textbf{TraceForge-123K dataset distribution.}
  Our corpus contains 1.8M observation–trace–language triplets, spanning tabletop, egocentric, and in-the-wild footage with moving cameras to support generalization across embodiments and scenes.}
  \label{fig:dataset_overview}
\end{figure}

\paragraph{Data engine: TraceForge.}
To enable scalable training, we introduce \traceforgelogo \textbf{TraceForge}, which consolidates heterogeneous sources---from controlled in-lab robot videos to in-the-wild human videos---into a unified 3D trace representation. TraceForge compensates camera motion, reconstructs frame-level trajectories from multiple viewpoints, and applies speed retargeting to normalize embodiment-dependent motion. The resulting dataset comprises 123K videos and 1.8M observation--trace--language triplets, providing diverse supervision for a robust 3D motion prior.

\paragraph{Results in low-data regimes.}
We evaluate two low-data adaptation settings that differ in the source and embodiment: \emph{(i) Robot$\rightarrow$Robot adaptation (small in-domain warm-up)}—with a five warm-up set of target-robot videos, TraceGen attains $80\%$ success across four tasks; and \emph{(ii) Human$\rightarrow$Robot transfer (no target-robot data)}—fine-tuning TraceGen only on five uncalibrated human demonstration videos recorded with a handheld phone in a different scene yields $67.5\%$ real-robot success. In both settings, trace-space inference is $50$--$600\times$ faster than state-of-the-art video-generation-based world models.

\paragraph{Contributions.}
Our contributions are:
\begin{itemize}
\item \tracegenlogo \textbf{TraceGen}: a world model that operates in 3D trace-space, enabling learning \emph{from} cross-embodiment, cross-environment, and cross-task videos by abstracting appearance and camera variation.
\item \traceforgelogo \textbf{TraceForge}: a unified pipeline that converts cross-embodiment videos into consistent 3D traces via camera-motion compensation, and speed retargeting.
\item %
\textbf{Scalable 3D trace learning and unified policy:} training on 1.8M observation--trace--language triplets across 123K videos ($>$15$\times$ prior work) to learn a \emph{single, embodiment-agnostic policy in trace space} that predicts scene-level 3D motion without detectors or heuristic filtering.
\item \textbf{Efficient few-shot adaptation}: $80\%$ success across four tasks with five in-domain robot videos and $67.5\%$ success from five human demos (human $\rightarrow$ robot transfer from handheld, uncalibrated camera), while achieving $>50\times$ faster inference than video-based world models. 
\end{itemize}

\section{Related Work}\label{sec:relatedwork}

\subsection{Embodied World Models}

\begin{figure}[t]
    \centering
    \includegraphics[width=1.\linewidth]{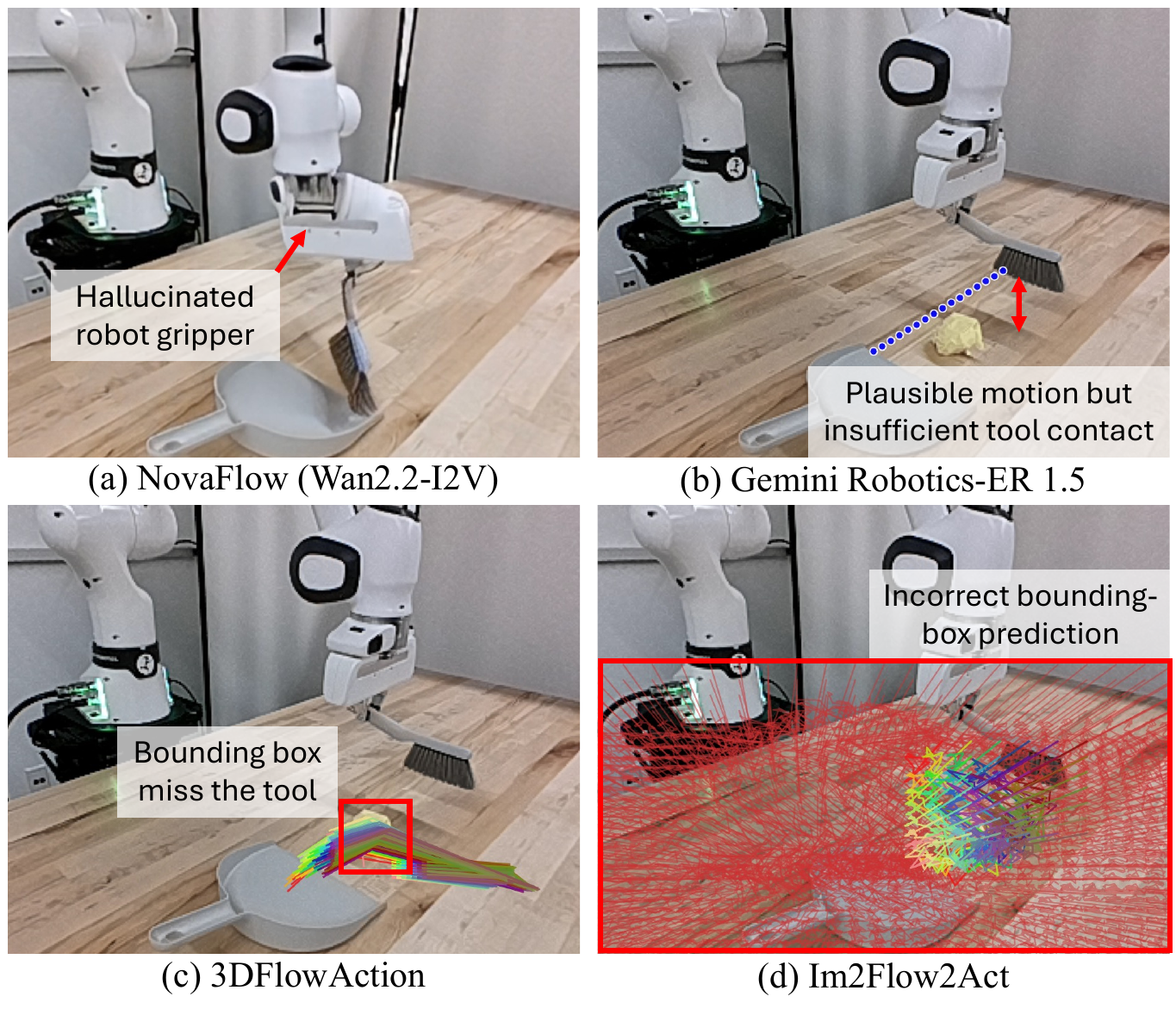}
    \vspace{-1.em}
    \caption{\textbf{Failure cases of existing embodied world models.}
(a) Video-based models can hallucinate geometry or affordance.
(b) VLM token outputs fail to capture fine motion. Bounding boxes miss the tool (c) or become overly broad (d).
}
    \label{fig:baseline_failures}
\end{figure}

World-model formulations for robotic manipulation span three major output-space families:

\textit{First}, video generation models predict raw pixels in future frames \cite{niu2025pre, zhen2025learning, li2025novaflow}. While expressive, they spend capacity reconstructing backgrounds and textures irrelevant to control, increasing computational cost, and risking geometry/affordance hallucinations (Fig. \ref{fig:baseline_failures}(a)).

\textit{Second}, language token-space models, such as VLM-based planners, generate discrete tokens; however, token-level outputs lack the spatial and temporal resolution required to represent fine object motion, limiting downstream control  \cite{lee2025molmoactactionreasoningmodels, li2025hamster, yang2025magma, abdolmaleki2025gemini} (Fig. \ref{fig:baseline_failures}(b)). 
Some works attempt to represent motions as skill tokens~\cite{collins2025amplify, kim2025uniskill}, but such representations inherit the limitations of their predefined extractors.

\textit{Third}, trace prediction models directly output future motion signals. Although more efficient and better aligned with control, previous work primarily trains on static, in-lab demonstrations and is largely restricted to 2D traces \cite{bharadhwaj2024track2act, xu2024flowcrossdomainmanipulationinterface, wen2023any, gao2024flip, ranasinghe2025langtomo}. Few 3D variants~\citep{zhi20253dflowaction} still focus solely on manipulated objects, requiring auxiliary object detection and heuristic filtering. These modules introduce error cascades and cannot capture robot motion, yielding an incomplete physical representation (Fig. \ref{fig:baseline_failures}(c)).

In contrast, \traceforgelogo \textbf{TraceForge} provides a lightweight pipeline that extends beyond in-lab data to in-the-wild videos, enabling the construction of large-scale training sets. Building on this, \tracegenlogo \textbf{TraceGen} is trained on 15$\times$ larger image–trace–language triple data than prior work~\citep{zhi20253dflowaction} and predicts scene-level 3D trajectories---robot and objects together---without heuristic filtering or bounding boxes. This yields a unified motion representation suitable for cross-embodiment learning.

\begin{figure*}[t!]
          \centering
          \includegraphics[width=1\linewidth]{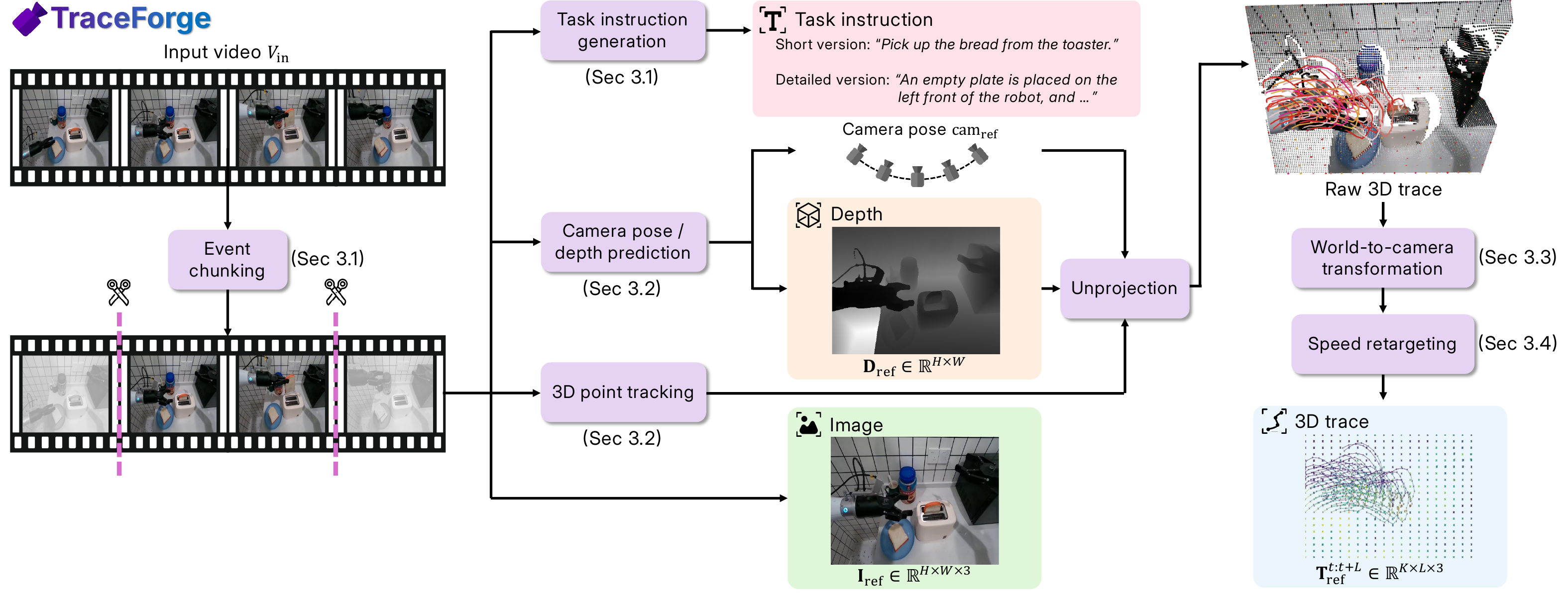}
          \vspace{-2.em}
          \caption{
          \textbf{Building the \traceforgelogo TraceForge dataset.}
            From an input video $V_{\mathrm{in}}$: (i) chunk task-relevant spans for curation and generate task instructions (\cref{subsec:chunking}); (ii) estimate camera pose and depth, select a reference image and track 3D points to form a raw trace (\cref{subsec:ptstracking}); (iii) apply world–to-camera alignment  (\cref{subsec:w2c}); (iv) speed retargeting to produce the final 3D trace (\cref{subsec:retarg}).}
          \label{fig:TraceForge}
\end{figure*}

\emph{Implicit world models for representation learning.}
\emph{Orthogonal} to the output-space families above, a body of work learns \emph{implicit} world models that shape latent dynamics for control without explicitly decoding future pixels or object/scene traces.  
\citep{Zhou2025Dino-wm, sun2023smart, zheng2023texttt, zheng2024premier, cui2024dynamo}. These methods have shown strong representation transfer, but typically operate in 2D feature space and do not provide metrically consistent, scene‑level 3D motion; precise object/end‑effector trajectories then require additional modules. TraceGen is complementary: it \emph{explicitly} models future motion in a compact 3D trace space, yielding a physically grounded, retargetable representation; in principle, implicit objectives can be layered onto TraceGen’s encoder to further strengthen pretraining.

\subsection{Trace for Robot Manipulation}

While scaling visual imitation has achieved promising manipulation skills, large models often need numerous expert robot trajectories and still struggle to generalize to new objects and scenes~\cite{finn2017one, mandlekar2018roboturk, young2021visual, zitkovich2023rt, bharadhwaj2024roboagent, liang2024make, jiang2024robots}. 
To improve transfer while reducing reliance on robot-only data, previous work leverages structured motion representations — trace. The predicted trace can be used by a variety of downstream modules,
such as planning or tracking-based execution~\citep{goyal2022ifor, eisner2022flowbot3d, seita2023toolflownet, vecerik2024robotap, li2025novaflow, yu2025genflowrl}, supervision or observation for policy learning~\citep{gu2023rt, zheng2024tracevla, yu2025sketch}, or high-level planning. In this work, we adopt a basic tracking controller as a minimal demonstration of executing our scene-level 3D traces; developing more sophisticated policies is left for future work.

\section{\traceforgelogo\hspace{0.1em} TraceForge: Dataset Construction}
\label{sec:dataset}

\paragraph{Overview of the TraceForge-TraceGen Pipeline.} TraceForge and TraceGen together form a unified world-modeling framework.
TraceForge serves as a scalable data engine (Sec. \ref{sec:dataset}), converting heterogeneous human and robot videos into consistent 3D trace annotations paired with multimodal observations and language. TraceGen (Sec. \ref{sec:training}) is trained on these large-scale trace-annotated triplets to learn a scene-level motion prior that predicts future trajectories directly in 3D trace-space. The next sections detail each component.

We introduce \traceforgelogo\textbf{TraceForge}, a unified pipeline that turns heterogeneous human and robot videos into large-scale, trace-annotated world-modeling data. Unlike prior work limited to static cameras or object-centric filtering, TraceForge operates directly on in-the-wild footage with moving viewpoints: it estimates camera pose, compensates camera motion, and reprojects traces into a fixed reference camera $\mathrm{cam}_{\mathrm{ref}}$. Each episode is paired with automatically generated task instructions, yielding multimodal triplets of \{observation, trace, language\}. Using TraceForge, we curate 123K episodes ($\sim$1.8M observation–trace–language triplets) from eight sources spanning human demonstrations, single-arm robot manipulation, and bimanual robot manipulation~\cite{goyal2017something, damen2020epic, mandlekar2021matters, liu2023libero, walke2023bridgedata, khazatsky2024droid, o2024open, bu2025agibot_iros, dass2025datamil}.

\subsection{Event Chunking and Instruction Generation}
\label{subsec:chunking}

 To align traces and language with underlying actions, we isolate task-related segments from each video and use them to construct observation–trace–language triplets. When start–end event indices are available, we extract the corresponding segment, splitting episodes with multiple labels into separate chunks. Otherwise, we identify task-relevant frames by removing those with negligible motion, as determined from point-tracking results.

For each event chunk, we generate diverse task instructions to better reflect how humans naturally specify goals to a robot and to reduce sensitivity to any single phrasing. Using a VLM, we produce three complementary instructions: (i) a short imperative, (ii) a multi-step decomposition, and (iii) a natural, human-like request. When the dataset already provides a human-written instruction, we keep it and augment it with these three variants. Otherwise, we sample representative frames from the start, middle, and end of the chunk and prompt the VLM to propose task instructions.
\begin{figure*}
      \centering
      \includegraphics[width=1\linewidth]{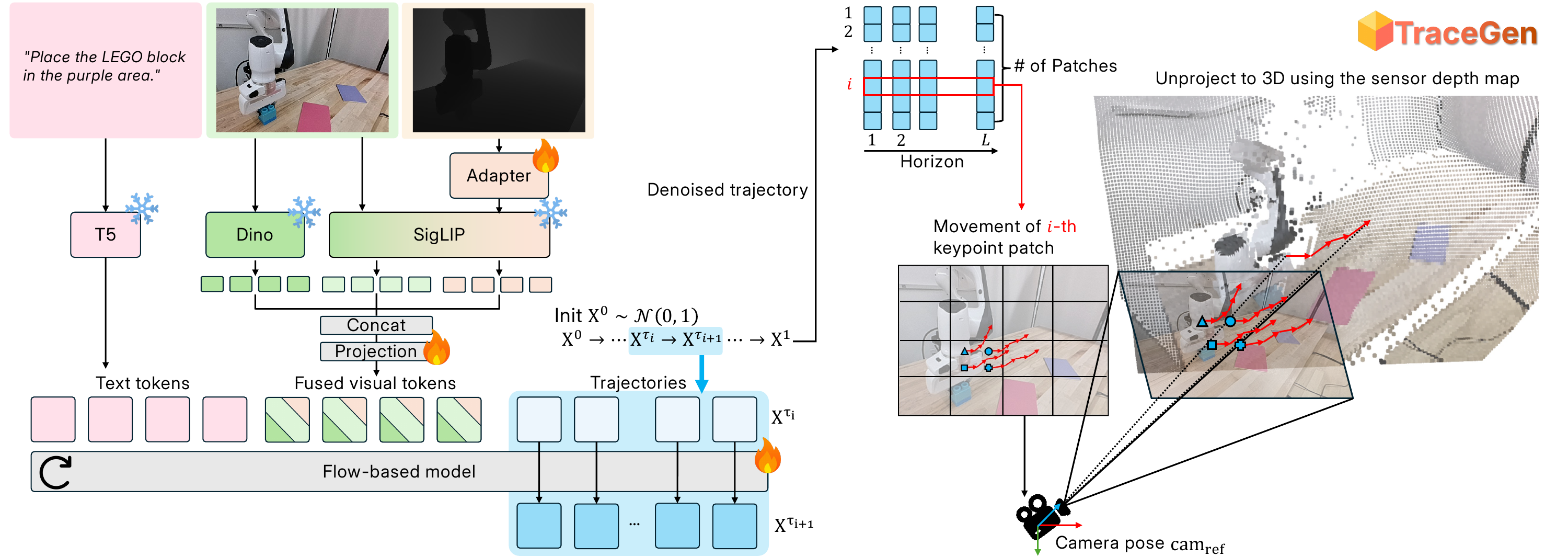}
            \vspace{-2.em}
            \caption{\textbf{Overview of \tracegenlogo TraceGen.} Given language, RGB, and depth inputs, text is encoded by a frozen T5 encoder, RGB images are processed by DINOv3 and SigLIP, and depth maps are passed through a SigLIP encoder with a learnable stem adapter. The resulting visual features (RGB + depth) are concatenated and linearly projected to form unified visual tokens. Together with text tokens, these serve as conditioning inputs to a CogVideoX-based flow model, which predicts a velocity field that transforms Gaussian noise into trace patches via ODE integration. 
            $\mathbf{X}^1$ represents the velocity-like 3D keypoint increments across frames predicted by the flow decoder, where $0, \cdots \tau_i, \tau_{i+1}, \cdots, 1$ denote the continuous interpolation times from pure noise to the clean trace increments.
            The predicted patches are then unpatched into 3D keypoint trajectories, expressed in the camera coordinate frame. These trajectories can be executed using various low-level controllers; in our experiments, we apply inverse kinematics to map predicted 3D traces to robot joint commands.}
      \label{fig:TraceGen}
  \end{figure*}

\subsection{3D Point Tracking with Camera Pose and Depth Prediction}
\label{subsec:ptstracking}
We extract 3D traces from videos with camera motion by recovering per-frame traces from each camera viewpoint. At the beginning of each event chunk, we select a reference frame, place a uniform \(20\times 20\) grid of keypoints \(K\) on its image, and track these points for a trace length of \(L\) steps. Instead of representing traces in full camera coordinates, we model each 3D trace point as \((x,y,z)\), where \((x,y)\) denotes the image-plane coordinates and \(z\) is the corresponding depth. This allows 3D traces to share the same screen alignment as 2D traces, enabling co-training and consistent supervision across both 2D and 3D modalities.

For 3D estimations of a video, including camera pose, depth, and 3D point traces, we adopt TAPIP3D~\cite{zhang2025tapip3dtrackingpointpersistent} as the 3D tracking model with CoTracker3~\cite{karaev2024cotracker3simplerbetterpoint, kumar2025trokens} as the point tracker. To improve efficiency, we replace its MegaSAM~\cite{li2025megasam} component with a fine-tuned VGGT~\cite{vggt} depth and camera pose predictor from SpatialTrackerV2~\cite{xiao2025spatialtrackerv2}, which achieves comparable accuracy while providing significantly faster inference without 3D optimization. Given an event chunk, our model generates per-frame camera poses and depth maps, and then reconstructs 3D point traces for the tracked keypoints. We designate the reference camera frame as \(\mathrm{cam}_{\mathrm{ref}}\) and its depth map as \(\mathbf{D}_{\mathrm{ref}}\), and express all 3D traces in the coordinate system of \(\mathrm{cam}_{\mathrm{ref}}\), providing a consistent reference frame that effectively compensates for camera motion during data curation. We additionally run CoTracker3 as a pure 2D point tracker on videos that require only image-plane motion, yielding extra 2D-only traces that increase the overall dataset size. Approximately \(20\%\) of all traces in our corpus are 2D-only.

\subsection{World-to-camera Transformation}
\label{subsec:w2c}
We transform all 3D traces to the reference camera frame $\mathrm{cam}_{\mathrm{ref}}$ to maintain point-of-view-consistency across time.
Given $K$ 3D traces in the world coordinates, we first use the estimated camera extrinsics at $\mathrm{cam}_{\mathrm{ref}}$ to transform them to camera coordinates, yielding $[X^c, Y^c, Z^c]^\top$.
Subsequently, we obtain the pixel coordinates of the traces, $(x, y)$, transformed by the estimated camera intrinsics.
Finally, we compose the pixel coordinates and depth values as screen-aligned 3D traces $\mathbf{T}_\mathrm{ref}^{t:t+L}=[x_i, y_i, z_i]^{t+L}_{i=t}$, where $L$ denotes the number of timesteps and $z=Z^c$.

\subsection{Speed Retargeting}
\label{subsec:retarg}
Human and robot demonstrations of the same task often differ in duration and execution speed. If we use these traces, the model sees the same behavior with different lengths and time scales, making it harder to learn a consistent motion representation. To make traces comparable across episodes and embodiments, we apply speed retargeting.

Each trace is temporally normalized to a fixed length \(L\) while preserving its relative motion profile. Specifically, we compute the cumulative arc length along the 3D path, reparameterize by normalized arc-length parameter, and resample at \(L\) uniformly spaced targets. This yields consistently sampled, training-ready traces that align in length across embodiments without distorting local velocity patterns.

\section{\tracegenlogo\hspace{0.1em} TraceGen: Architecture and Training}

\label{sec:training}

We present \tracegenlogo\textbf{TraceGen}, a flow-based world model that predicts future 3D motion trajectories from multimodal observations. Our model builds on the CogVideoX~\cite{yang2024cogvideox} architecture and employs a Prismatic-VLM~\cite{karamcheti2024prismatic} multi-encoder fusion strategy to integrate heterogeneous visual and linguistic information.

\subsection{Multi-Encoder Feature Extraction}

\paragraph{RGB encoders.}
We adopt a multi-stream encoding strategy that captures complementary visual representations. For each RGB input image $\mathbf{I} \in \mathbb{R}^{H \times W \times 3}$, we extract features using two frozen pretrained encoders:

\begin{itemize}[leftmargin=*,noitemsep,topsep=0pt]
    \item \textbf{DINOv3}~\cite{simeoni2025dinov3}: A self-supervised vision transformer (ViT-L/16) that produces spatially-aware geometric features $\mathbf{F}_{\text{dino}} \in \mathbb{R}^{N \times D_d}$.
    \item \textbf{SigLIP}~\cite{zhai2023sigmoid}: A vision-language model (SigLIP-Base-Patch16-384) that generates semantically aligned features $\mathbf{F}_{\text{siglip}} \in \mathbb{R}^{N \times D_s}$ suitable for text-conditioned prediction.
\end{itemize}
\begin{figure*}[t!]
  \centering
  \includegraphics[width=1.0\linewidth]{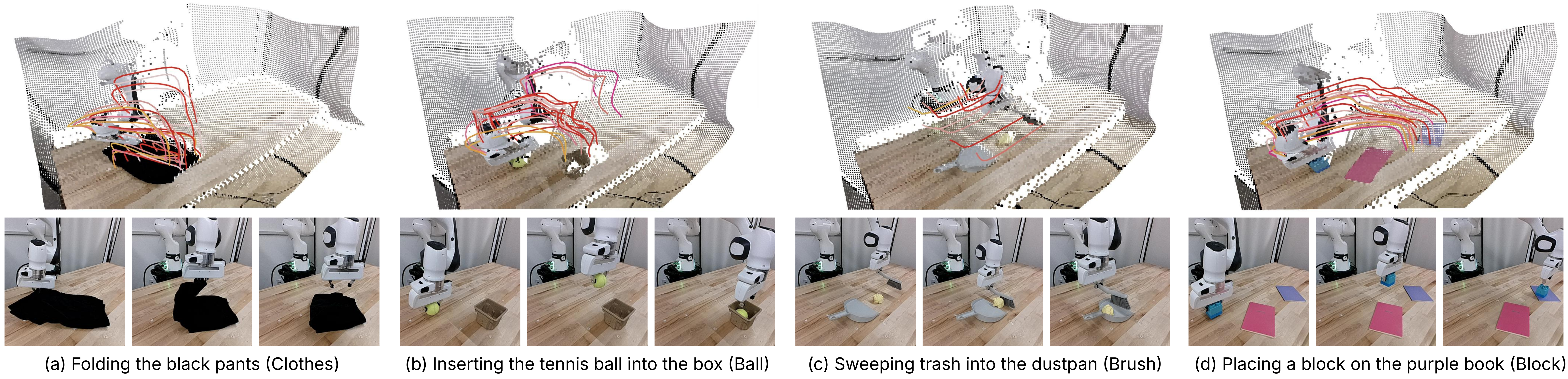}
    \vspace{-2.em}
    \caption{
\textbf{Real-world experiments with predicted 3D traces.} 
We evaluate TraceGen and baselines on four real-world manipulation tasks on a Franka Research 3 robot, showing that the predicted 3D traces transfer effectively to real-robot execution.
}
  \label{fig:real_world_setup_overview}
\end{figure*}
\paragraph{Depth encoder.}
To incorporate 3D geometric information, we process depth maps $\mathbf{D} \in \mathbb{R}^{H \times W}$ %
through a third encoder equipped with a learnable \textit{stem adapter}---a 1×1 convolutional layer that projects single-channel depth to the 3-channel input space expected by SigLIP, yielding $\mathbf{F}_{\mathrm{depth}}\in\mathbb{R}^{N\times D_s}$

\paragraph{Text encoder.}
Task instructions are encoded using a frozen T5-base~\cite{raffel2020exploring} encoder, 
producing contextualized text embeddings $\mathbf{F}_{\text{text}} \in \mathbb{R}^{M \times D}$, where we fix the text sequence length to $M = 128$ tokens and token dimension $D = 768$.

\paragraph{Prismatic VLM fusion.}
Following Prismatic VLM~\cite{karamcheti2024prismatic}, we concatenate the three vision streams \textit{along the feature dimension}:
\begin{equation}
\mathbf{F}_{\text{vis}} = \text{Concat}(\mathbf{F}_{\text{dino}}, \mathbf{F}_{\text{siglip}}, \mathbf{F}_{\text{depth}}) \in \mathbb{R}^{N \times (D_d + D_s + D_s)},
\end{equation}
then project to a unified dimension $D = 768$ via a learnable linear layer:
\begin{equation}
\mathbf{F}_{\text{vis}} = \text{Linear}(\mathbf{F}_{\text{vis}}) \in \mathbb{R}^{N \times D}.
\end{equation}
The visual tokens $\mathbf{F}_{\text{vis}} \in \mathbb{R}^{N \times D}$ and text tokens $\mathbf{F}_{\text{text}} \in \mathbb{R}^{M \times D}$ are combined to form the conditioning input $\mathbf{F}_{\text{cond}} \in \mathbb{R}^{(N + M) \times D}$ for the flow-based trace decoder.

\subsection{Flow-based Trace Decoder}

\paragraph{Architecture.}
Our decoder adapts CogVideoX's~\cite{yang2024cogvideox} 3D transformer to operate in trace space. The input is a $K \times L$ grid where $K = 20 \times 20$ spatial keypoints are tracked across $L = 32$ future timesteps, with each point as $(x, y, z) \in \mathbb{R}^3$ in the camera frame. We apply spatial patchification with patch size $2 \times 2$, where each $2 \times 2$ group of keypoints is processed as a single token, resulting in $10 \times 10$ spatial tokens per timestep. Following CogVideoX, we inject $\mathbf{F}_{\text{cond}}$ via Adaptive LayerNorms applied separately to contextual input and latent trace tokens, enabling efficient fusion.

\paragraph{Trace generation via stochastic interpolants.}

Our model aims to generate the 3D trace of the scene, denoted as $\mathbf{T}_\mathrm{ref}^{t:t+L}$. Each $\mathbf{T}_\mathrm{ref}^{t}$ corresponds to a $20 \times 20$ uniform grid with depth map value at time $t$. Thus, instead of predicting these absolute grid values directly, we observe that the full 3D trace $\mathbf{T}_\mathrm{ref}^{t:t+L}$ can be equivalently reconstructed from the temporal differences 

\begin{equation}
    \Delta\mathbf{T}_{\mathrm{ref}}^{t}
    = \mathbf{T}_{\mathrm{ref}}^{t+1} - \mathbf{T}_{\mathrm{ref}}^{t}.
\end{equation}

Therefore, our neural network is trained to predict velocity-like increments in keypoints, implicitly capturing the scene's underlying 3D motion.

We adopt the Stochastic Interpolant framework~\cite{albergo2023stochastic}, which unifies diffusion-based and flow-based generative models by defining an interpolation path between data and noise distributions. To streamline notation, we denote the keypoint increments $\Delta\mathbf{T}_{\mathrm{ref}}^{t}$ as $\mathbf{X} \in \mathbb{R}^{K \times L \times 3}$, which serves as our target data distribution. The framework introduces a stochastic interpolant:
\begin{equation}
\mathbf{I}_\tau = \alpha_\tau \mathbf{X}^1 + \sigma_\tau \boldsymbol{\varepsilon}, \quad \tau \in [0, 1],
\label{eq:interpolant}
\end{equation} 
where $\mathbf{X}^1$ is the ground-truth trace, $\boldsymbol{\varepsilon} \sim \mathcal{N}(0, \mathbf{I})$ is Gaussian noise, and $\alpha_\tau$, $\sigma_\tau$ are time-dependent schedules. By varying $\alpha_\tau$ and $\sigma_\tau$, this framework encompasses a range of generative models, including diffusion models and flow-matching methods.

The framework learns a velocity field $\mathbf{v}(\mathbf{x}, \tau, \mathbf{F}_{\text{cond}})$ that characterizes the time evolution of the interpolant, where $\mathbf{x}$ denotes a sample from the interpolant distribution at time $\tau$:
\begin{equation}
\mathbf{v}(\mathbf{x}, \tau, \mathbf{F}_{\text{cond}}) = \mathbb{E}[\dot{\mathbf{I}}_\tau \mid \mathbf{I}_\tau = \mathbf{x}, \mathbf{F}_{\text{cond}}],
\end{equation}
where $\dot{\mathbf{I}}_\tau$ denotes the time derivative and the expectation is over $\mathbf{X}^0, \mathbf{X}^1$ conditioned on $\mathbf{I}_\tau = \mathbf{x}$ and $\mathbf{F}_{\text{cond}}$.

\paragraph{Linear interpolation ODE.}
Among the variants within the Stochastic Interpolant framework, we implement a \textit{linear interpolation ODE} by choosing $\alpha_\tau = \tau$ and $\sigma_\tau = 1-\tau$:
\begin{equation}
\mathbf{X}^\tau = (1 - \tau) \mathbf{X}^0 + \tau \mathbf{X}^1, \quad \tau \in [0, 1],
\end{equation}

With this linear schedule, the velocity field simplifies to $\dot{\mathbf{X}}^\tau = \mathbf{X}^1 - \mathbf{X}^0$, which is constant in time. We train a neural network $v_\theta$ to predict this velocity by minimizing:
\begin{equation}
\mathcal{L}_{\text{SI}} = \mathbb{E}_{\tau, \mathbf{X}^0, \mathbf{X}^1} \left[ \| v_\theta(\mathbf{X}^\tau, \tau, \mathbf{F}_{\text{cond}}) - (\mathbf{X}^1 - \mathbf{X}^0) \|^2 \right].
\end{equation}

At test time, we generate trajectories via 100-step ODE integration, relying solely on the conditional model with multimodal vision-language conditioning.

\paragraph{Encoder freezing strategy.}
To leverage pretrained representations efficiently, we keep all encoders 
(DINOv3, SigLIP, T5) frozen throughout training and trains only the fusion 
layer and decoder, following \cite{karamcheti2024prismatic}.

\begin{figure*}[t!]
  \centering
  \includegraphics[width=1.\linewidth]{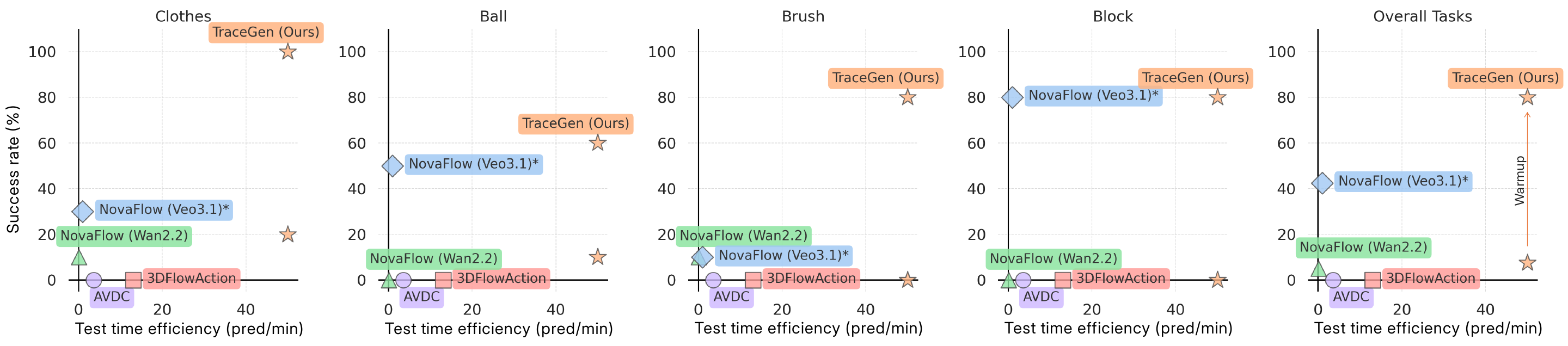}
    \vspace{-2.em}
    \caption{
    \textbf{Success rate vs.\ inference efficiency} (predictions per minute; higher and rightward is better). \tracegenlogo \textbf{TraceGen} achieves the best combination of success and efficiency, outperforming both video and trace-based baselines by a large margin. Gains stem from its strong 3D motion prior and a \emph{lightweight warm-up} in trace space via \traceforgelogo \textbf{TraceForge}. 
    In contrast, video-generation baselines (e.g., NovaFlow or video backbone in AVDC) offer no practical few-shot warm-up path in our setting, and several trace baselines rely on object detectors or heuristic object filtering, making warm-up technically difficult. (The Veo~3.1 latency is measured based on its average API call time.)}
    
  \label{fig:main_result}
\end{figure*}

\section{Experiments}
\label{sec:experiments}

Our experiments address three questions:
(1) \textbf{Effectiveness of TraceGen}: Does planning in compact 3D trace space improve performance and inference efficiency compared to pixel-based alternatives? (\cref{sec:trace_gen_performance})
(2) \textbf{Human--Robot Transfer}: Can TraceGen enable efficient human-to-robot transfer from uncalibrated, in-the-wild videos with differing camera, backgrounds, and object layouts? (Sec.~\ref{sec:human_robot_adaptation})
(3) \textbf{Role of Pretraining and Warmup}: How much do large-scale cross-embodiment pretraining and lightweight warmup contribute to performance and 
generalization? (Sec.~\ref{sec:ablation_study}) (We report a quantitative sanity check of TraceForge trace accuracy in the Appendix \ref{traceforge_result}.)

\paragraph{Warm-up rationale.}
\tracegenlogo \textbf{TraceGen} learns a unified policy in 3D trace space, predicting future scene-level trajectories that are embodiment-agnostic. To execute on a specific robot, these traces must be ``\emph{translated}'' into the robot’s action space via a lightweight warm-up.

\noindent\textbf{Settings.}
We evaluate two lightweight regimes that differ only in the source of supervision for warm-up (For both warm-up regimes, we do not cherry-pick demonstrations. We also provide visualizations of \emph{all} warm-up data in the Appendix \ref{warmup_data}.):
\begin{enumerate}[leftmargin=*, itemsep=2pt, topsep=2pt]
\item \textbf{Robot$\rightarrow$Robot (small same-embodiment warm-up).}
We fine-tune \tracegenlogo \textbf{TraceGen} using a \textbf{five} in-domain set of robot demonstrations.
The demonstrations differ from the target tasks in the object/target configuration and the robot’s initial pose. For example, in \emph{Brush}, demonstrations begin with the brush already in contact with (or very close to) the table and thus omit the critical ``lower the brush'' motion required at test time. Similarly, for \emph{Block}, demonstration videos use target regions with randomly varying colors and positions.

\item \textbf{Human$\rightarrow$Robot (no target-robot data).}
We fine-tune \tracegenlogo \textbf{TraceGen} with \textbf{five} uncalibrated human demonstrations (handheld phone, different scene) to adapt trace predictions to the target task; no target-robot demonstrations are used.
Each video is only \textbf{3–4 seconds} long, and a single person performing the task while another records it is sufficient to obtain the data. Overall, collecting \textbf{20 demonstrations} across four tasks required \textbf{under 4 minutes}, making the warm-up extremely easy.
\end{enumerate}

\subsection{Performance and Efficiency Comparison in Real-World Experiment}
\label{sec:trace_gen_performance}

\paragraph{Tasks and setup.}
We evaluate on four manipulation tasks executed on a Franka Research 3 robot: folding a garment (\emph{Clothes}), inserting a tennis ball into a box (\emph{Ball}), sweeping trash into a dustpan with a brush (\emph{Brush}), placing a block in the purple region (\emph{Block}).
Given a single RGB-D frame and a language instruction, TraceGen predicts a 3D trace, which is converted to joint commands via inverse kinematics.

\paragraph{Baselines.}
We include both video-based and trace-based world models. Video-generation approaches such as AVDC~\cite{ko2023learning} and NovaFlow~\cite{li2025novaflow} first synthesize future video and then estimate 3D motion post hoc; consistent with NovaFlow’s evaluation, we use only the video-generation component of AVDC and apply a unified video-to-trace extraction pipeline across all video-based baselines. For 3DFlowAction~\cite{zhi20253dflowaction}, which relies on segmentation masks, we supply ground-truth masks due to frequent failures of the original mask estimator.

\paragraph{Performance.}
Fig.~\ref{fig:main_result} shows that all methods below 10B parameters---\textbf{except \tracegenlogo TraceGen (0.67B)}---fail to produce executable trajectories in the zero-shot setting (0\% success across all tasks).
Large video-generation models---NovaFlow (Wan2.2) and NovaFlow (Veo3.1)---achieve non-zero zero-shot success, but at the cost of extremely slow inference. Under the same 5-video warm-up procedure, \tracegenlogo \textbf{TraceGen} attains \textbf{80\%} success across four tasks despite variations in object layouts and initial robot poses. Large video-generation models exceeding 10B parameters are impractical to warm up due to proprietary APIs or substantial computational requirements. These results suggest that TraceGen’s few-shot adaptability stems from its compact trace-space representation and the TraceForge pipeline, which together provide stable motion priors and consistent supervision for lightweight warm-up.

\paragraph{Inference efficiency.}
Planning in 3D trace space offers substantial computational benefits. \tracegenlogo \textbf{TraceGen} runs $3.8\times$ faster than trace-generation baselines and over \textbf{$50\times$} faster than large video-generation models. NovaFlow (Wan2.2) requires more than \textbf{$600\times$} longer inference time, highlighting the difficulty of scaling pixel-space video prediction for real-time robotics. TraceGen thus provides a practical and efficient solution for closed-loop planning.

\subsection{Human--Robot Skill Transfer}
\label{sec:human_robot_adaptation}

\begin{figure}[h!]
  \centering
  \includegraphics[width=.9\linewidth]{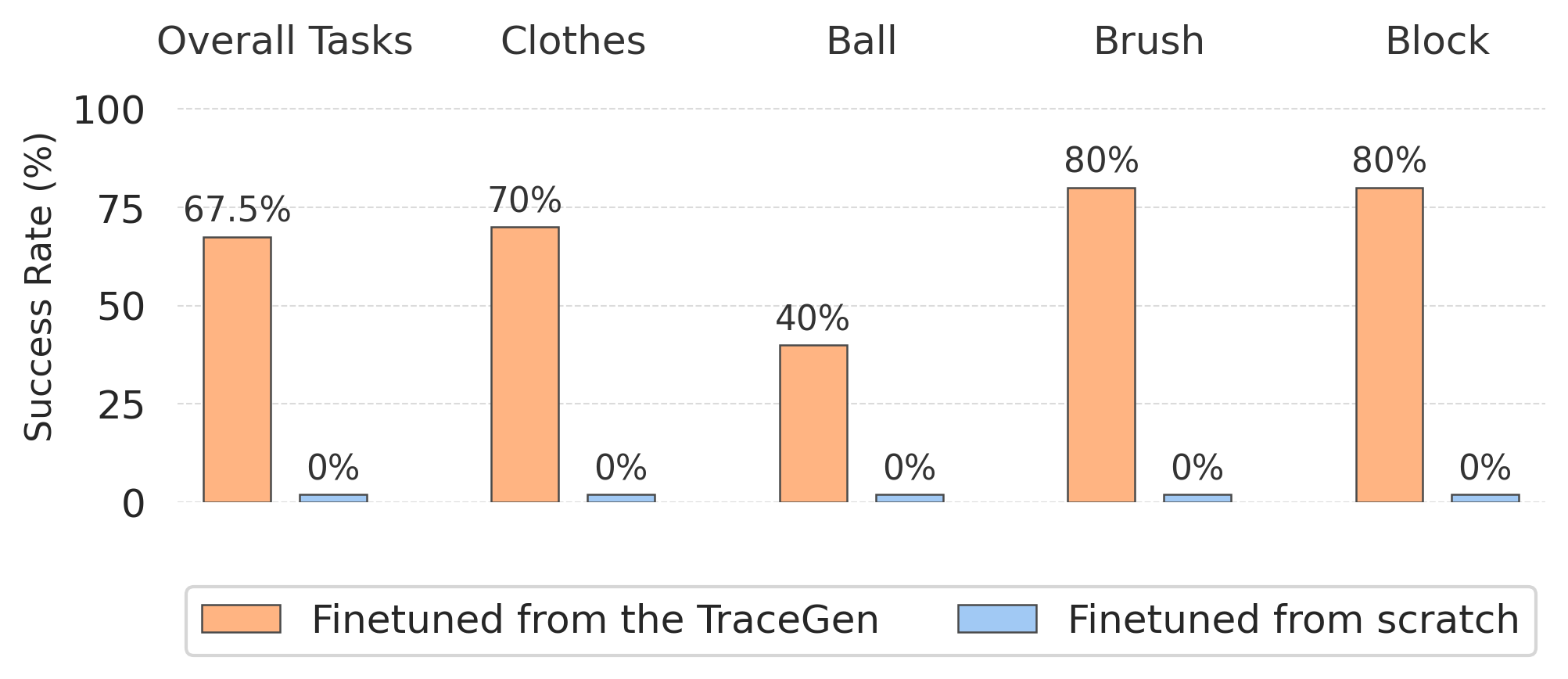}
  \vspace{-1.2em}
  \caption{\textbf{Human-to-robot skill transfer using human demo videos.}
  TraceGen, finetuned on 5 in-the-wild handheld phone videos, successfully executes four manipulation tasks, with a success rate of 67.5\%.
  In contrast, the From Scratch model fails (0\%), indicating that cross-embodiment pretraining is essential.}
  \label{fig:human_to_robot}
  \vspace{-1.7em}
\end{figure}

\paragraph{Protocol and results.}
We evaluate whether TraceGen can transfer skills from in-the-wild human demo videos to a real robot. For each task, we collect five handheld phone videos recorded without camera calibration, with varying viewpoints, backgrounds, and object layouts. TraceForge reconstructs 3D traces from these demonstrations, and TraceGen is finetuned on the resulting traces before deployment on a Franka Research 3 robot. As shown in Fig. \ref{fig:human_to_robot}, finetuning on the five human demos yields an overall success rate of \textbf{67.5\%} across four tasks, whereas the \emph{From Scratch} model fails on all tasks (0\%). Despite substantial differences in embodiment, camera intrinsics, and scene appearance, the pretrained TraceGen model adapts effectively with only a small number of uncalibrated human videos, indicating that the 3D trace representation provides a practical bridge between human demonstrations and robot execution.

\subsection{Role of Pretraining and Warmup}
\label{sec:ablation_study}

We investigate how large-scale cross-embodiment pretraining affects TraceGen's ability to adapt with few task demonstrations. We compare the pretrained model with a \textit{From Scratch} variant that shares the same architecture but is trained only on warmup data.

\paragraph{Few-shot warmup and the importance of pretraining.}
Table~\ref{tab:task_comparison_transposed} summarizes the effect of 5-video and 15-video warmups. With five target-robot videos, the pretrained model achieves an overall success rate of \textbf{80\%}, whereas the scratch model attains \textbf{25\%}. Increasing warmup to fifteen videos yields limited additional improvement for the pretrained model (\textbf{82.5\%}) and marginal change for the scratch variant (\textbf{30\%}). These results indicate that the majority of TraceGen’s performance stems from pretraining, with warmup primarily aligning pretrained motion priors with task-specific configurations.

\begin{table}[ht]
\centering
\caption{
Effect of cross-embodiment pretraining under 5-video and 15-video warmup.
Pretraining significantly improves success rates compared to training from scratch.
}
\label{tab:task_comparison_transposed}
\resizebox{\columnwidth}{!}{%
\begin{tabular}{clccccc}
\toprule
\textbf{Warm-up} & \textbf{Pretraining} & \textbf{Clothes} & \textbf{Ball} & \textbf{Brush} & \textbf{Block} & \textbf{Overall SR(\%)} \\
\midrule

\multirow{2}{*}{\textbf{5 robot videos}} 
 & \textbf{Random init.} & 10/10& 0/10 & 0/10 & 0/10 & \textbf{25.0\%} \\
 & \textbf{TraceGen}     & 10/10 & 6/10 &8/10 & 8/10 & \textbf{80\%} \\
\midrule

\multirow{2}{*}{\textbf{15 robot videos}} 
 & \textbf{Random init.} & 10/10 & 0/10 & 0/10 & 2/10  & \textbf{30.0\%} \\
 & \textbf{TraceGen}      & 10/10 & 9/10  & 8/10 & 6/10  & \textbf{82.5\%} \\

\bottomrule

\end{tabular}
}
\end{table}

\paragraph{Effect of pretraining source.}
To quantify the role of pretraining data, we compare four variants trained with the same 5-video warmup set but different pretraining sources: (1) no pretraining (\emph{From Scratch}); (2) pretraining on SSV2 (human hand–centric, 35K clips); (3) pretraining on Agibot (robot‑centric, 35K clips); and (4) full TraceGen pretraining on the cross-embodiment dataset. Under identical warmup, SSV2 pretraining yields \textbf{25\%} success and Agibot yields \textbf{45\%}, both lower than cross-embodiment pretraining on the full dataset. These results suggest that both embodiment alignment (robot-centric data) and heterogeneous motion coverage (human + robot sources) matter, and combining them yields substantially better transfer.
\begin{table}[h]
\centering
\caption{Effect of pretraining source on 5-video warmup performance.
Cross-embodiment pretraining with a larger dataset (TraceGen) yields substantially higher success than single-source pretraining and full scratch training.}
\label{tab:task_comparison}
\resizebox{\columnwidth}{!}{%
\begin{tabular}{lcccc}
\toprule
\textbf{Task} & \textbf{From scratch} & \textbf{SSV2 only} &  \textbf{Agibot only} & \textbf{TraceForge-123K} \\
\midrule
\textbf{Ball} & 0/10 & 3/10 & 4/10 & 6/10 \\
\textbf{Block} & 0/10 & 2/10 & 5/10 & 8/10 \\
\hline
\textbf{Overall SR(\%)} & 0\% & 25\% & 45\%& 70\% \\
\bottomrule
\end{tabular}
}
\end{table}

\section{Conclusion}
\label{sec:conclusion}
We presented \tracegenlogo \textbf{TraceGen}, a cross-embodiment world model that predicts future motion in compact 3D trace space rather than pixel space. By representing manipulation tasks as 3D traces of scene points, TraceGen achieves a unified motion representation that generalizes across diverse embodiments—from human hands to robot arms. To enable large-scale training, we introduced \traceforgelogo \textbf{TraceForge}, a data-curation pipeline that processes heterogeneous sources into consistent 3D traces by compensating for camera motion and normalizing embodiment-specific speeds. Pretrained on 123K episodes, TraceGen achieves 80–82.5\% success on real-world tasks with only 5–15 demonstrations, running 50× faster than video-based approaches. These results suggest that reasoning in trace space provides an effective inductive bias for cross-embodiment learning, offering both computational efficiency and sample efficiency for robot manipulation.

\section{Limitations and Future Work}
\label{sec:future_work}

Within the Stochastic Interpolant framework, we adopt linear interpolation with ODE integration. While this approach allows sampling diverse trajectories through different noise initializations, we have not yet explored alternative interpolation schedules or mechanisms to explicitly control which trajectory mode is generated for ambiguous tasks.

The quality of demonstration data varies. A portion of our source videos contains inefficient or corrective motions—where operators make exploratory movements or errors before completing tasks—introducing suboptimal supervision signals. We implemented additional filtering steps to clean the dataset, though some noisy demonstrations remain.

Moreover, TraceGen's zero-shot generation ability, while promising, is not yet fully reliable under novel embodiments or unseen environments, occasionally yielding plausible but physically infeasible trajectories. Additionally, for fine-grained manipulation tasks, the generated trajectories may lack sufficient detail for the robot to execute precise manipulation actions. Scaling to internet-scale demonstration datasets, combined with improved data filtering mechanisms, could address these issues. Finally, extending beyond human-like robot arms to very different robot types would test the limits of trace-space abstraction. Despite these challenges, we believe TraceGen's efficiency and generality represent a meaningful step toward practical cross-embodiment manipulation systems.

\section*{Acknowledgements}
Seungjae Lee, Yoonkyo Jung, Zikui Cai, Yongyuan Liang, and Furong Huang are supported by DARPA Transfer from Imprecise and Abstract Models to Autonomous Technologies (TIAMAT) 80321, DARPA HR001124S0029-AIQ-FP-019, DOD-AFOSR-Air Force Office of Scientific Research under award number FA9550-23-1-0048, National Science Foundation NSF-IIS-2147276 FAI, National Science Foundation NAIRR240045, National Science Foundation TRAILS Institute (2229885). Private support was provided by Peraton and Open Philanthropy. We thank Shivin Dass and Daniel Ekpo for their help and advice.

{
    \small
    \bibliographystyle{ieeenat_fullname}
    \bibliography{main}

\begin{thebibliography}{64}
\providecommand{\natexlab}[1]{#1}
\providecommand{\url}[1]{\texttt{#1}}
\expandafter\ifx\csname urlstyle\endcsname\relax
  \providecommand{\doi}[1]{doi: #1}\else
  \providecommand{\doi}{doi: \begingroup \urlstyle{rm}\Url}\fi

\bibitem[Abdolmaleki et~al.(2025)Abdolmaleki, Abeyruwan, Ainslie, Alayrac, Arenas, Balakrishna, Batchelor, Bewley, Bingham, Bloesch, et~al.]{abdolmaleki2025gemini}
Abbas Abdolmaleki, Saminda Abeyruwan, Joshua Ainslie, Jean-Baptiste Alayrac, Montserrat~Gonzalez Arenas, Ashwin Balakrishna, Nathan Batchelor, Alex Bewley, Jeff Bingham, Michael Bloesch, et~al.
\newblock Gemini robotics 1.5: Pushing the frontier of generalist robots with advanced embodied reasoning, thinking, and motion transfer.
\newblock \emph{arXiv preprint arXiv:2510.03342}, 2025.

\bibitem[Albergo et~al.(2023)Albergo, Boffi, and Vanden-Eijnden]{albergo2023stochastic}
Michael~S Albergo, Nicholas~M Boffi, and Eric Vanden-Eijnden.
\newblock Stochastic interpolants: A unifying framework for flows and diffusions.
\newblock \emph{arXiv preprint arXiv:2303.08797}, 2023.

\bibitem[Barreiros et~al.(2025)Barreiros, Beaulieu, Bhat, Cory, Cousineau, Dai, Fang, Hashimoto, Irshad, Itkina, et~al.]{barreiros2025careful}
Jose Barreiros, Andrew Beaulieu, Aditya Bhat, Rick Cory, Eric Cousineau, Hongkai Dai, Ching-Hsin Fang, Kunimatsu Hashimoto, Muhammad~Zubair Irshad, Masha Itkina, et~al.
\newblock A careful examination of large behavior models for multitask dexterous manipulation.
\newblock \emph{arXiv preprint arXiv:2507.05331}, 2025.

\bibitem[Bharadhwaj et~al.(2024{\natexlab{a}})Bharadhwaj, Mottaghi, Gupta, and Tulsiani]{bharadhwaj2024track2act}
Homanga Bharadhwaj, Roozbeh Mottaghi, Abhinav Gupta, and Shubham Tulsiani.
\newblock Track2act: Predicting point tracks from internet videos enables generalizable robot manipulation.
\newblock In \emph{ECCV}, 2024{\natexlab{a}}.

\bibitem[Bharadhwaj et~al.(2024{\natexlab{b}})Bharadhwaj, Vakil, Sharma, Gupta, Tulsiani, and Kumar]{bharadhwaj2024roboagent}
Homanga Bharadhwaj, Jay Vakil, Mohit Sharma, Abhinav Gupta, Shubham Tulsiani, and Vikash Kumar.
\newblock Roboagent: Generalization and efficiency in robot manipulation via semantic augmentations and action chunking.
\newblock In \emph{ICRA}, 2024{\natexlab{b}}.

\bibitem[Bjorck et~al.(2025)Bjorck, Casta{\~n}eda, Cherniadev, Da, Ding, Fan, Fang, Fox, Hu, Huang, et~al.]{bjorck2025gr00t}
Johan Bjorck, Fernando Casta{\~n}eda, Nikita Cherniadev, Xingye Da, Runyu Ding, Linxi Fan, Yu Fang, Dieter Fox, Fengyuan Hu, Spencer Huang, et~al.
\newblock Gr00t n1: An open foundation model for generalist humanoid robots.
\newblock \emph{arXiv preprint arXiv:2503.14734}, 2025.

\bibitem[Bu et~al.(2025)Bu, Cai, Chen, Cui, Ding, Feng, He, Huang, et~al.]{bu2025agibot_iros}
Qingwen Bu, Jisong Cai, Li Chen, Xiuqi Cui, Yan Ding, Siyuan Feng, Xindong He, Xu Huang, et~al.
\newblock Agibot world colosseo: A large-scale manipulation platform for scalable and intelligent embodied systems.
\newblock In \emph{IROS}, 2025.

\bibitem[Collins et~al.(2025)Collins, Cheng, Aneja, Wilcox, Joffe, and Garg]{collins2025amplify}
Jeremy~A Collins, Lor{\'a}nd Cheng, Kunal Aneja, Albert Wilcox, Benjamin Joffe, and Animesh Garg.
\newblock Amplify: Actionless motion priors for robot learning from videos.
\newblock \emph{arXiv preprint arXiv:2506.14198}, 2025.

\bibitem[Cui et~al.(2024)Cui, Pan, Iyer, Haldar, and Pinto]{cui2024dynamo}
Zichen Cui, Hengkai Pan, Aadhithya Iyer, Siddhant Haldar, and Lerrel Pinto.
\newblock Dynamo: In-domain dynamics pretraining for visuo-motor control.
\newblock In \emph{NeurIPS}, 2024.

\bibitem[Damen et~al.(2020)Damen, Doughty, Farinella, Fidler, Furnari, Kazakos, Moltisanti, Munro, Perrett, Price, et~al.]{damen2020epic}
Dima Damen, Hazel Doughty, Giovanni~Maria Farinella, Sanja Fidler, Antonino Furnari, Evangelos Kazakos, Davide Moltisanti, Jonathan Munro, Toby Perrett, Will Price, et~al.
\newblock The epic-kitchens dataset: Collection, challenges and baselines.
\newblock \emph{IEEE Transactions on Pattern Analysis and Machine Intelligence}, 43\penalty0 (11):\penalty0 4125--4141, 2020.

\bibitem[Dass et~al.(2025)Dass, Khaddaj, Engstrom, Madry, Ilyas, and Mart{\'\i}n-Mart{\'\i}n]{dass2025datamil}
Shivin Dass, Alaa Khaddaj, Logan Engstrom, Aleksander Madry, Andrew Ilyas, and Roberto Mart{\'\i}n-Mart{\'\i}n.
\newblock Datamil: Selecting data for robot imitation learning with datamodels.
\newblock \emph{arXiv preprint arXiv:2505.09603}, 2025.

\bibitem[Eisner et~al.(2022)Eisner, Zhang, and Held]{eisner2022flowbot3d}
Ben Eisner, Harry Zhang, and David Held.
\newblock Flowbot3d: Learning 3d articulation flow to manipulate articulated objects.
\newblock In \emph{RSS}, 2022.

\bibitem[Finn et~al.(2017)Finn, Yu, Zhang, Abbeel, and Levine]{finn2017one}
Chelsea Finn, Tianhe Yu, Tianhao Zhang, Pieter Abbeel, and Sergey Levine.
\newblock One-shot visual imitation learning via meta-learning.
\newblock In \emph{CoRL}, 2017.

\bibitem[Gao et~al.(2025)Gao, Zhang, Xu, Zhehao, and Shao]{gao2024flip}
Chongkai Gao, Haozhuo Zhang, Zhixuan Xu, Cai Zhehao, and Lin Shao.
\newblock Flip: Flow-centric generative planning as general-purpose manipulation world model.
\newblock In \emph{ICLR}, 2025.

\bibitem[Goyal et~al.(2022)Goyal, Mousavian, Paxton, Chao, Okorn, Deng, and Fox]{goyal2022ifor}
Ankit Goyal, Arsalan Mousavian, Chris Paxton, Yu-Wei Chao, Brian Okorn, Jia Deng, and Dieter Fox.
\newblock Ifor: Iterative flow minimization for robotic object rearrangement.
\newblock In \emph{CVPR}, 2022.

\bibitem[Goyal et~al.(2017)Goyal, Ebrahimi~Kahou, Michalski, Materzynska, Westphal, Kim, Haenel, Fruend, Yianilos, Mueller-Freitag, et~al.]{goyal2017something}
Raghav Goyal, Samira Ebrahimi~Kahou, Vincent Michalski, Joanna Materzynska, Susanne Westphal, Heuna Kim, Valentin Haenel, Ingo Fruend, Peter Yianilos, Moritz Mueller-Freitag, et~al.
\newblock The" something something" video database for learning and evaluating visual common sense.
\newblock In \emph{ICCV}, 2017.

\bibitem[Gu et~al.(2024)Gu, Kirmani, Wohlhart, Lu, Arenas, Rao, Yu, Fu, Gopalakrishnan, Xu, et~al.]{gu2023rt}
Jiayuan Gu, Sean Kirmani, Paul Wohlhart, Yao Lu, Montserrat~Gonzalez Arenas, Kanishka Rao, Wenhao Yu, Chuyuan Fu, Keerthana Gopalakrishnan, Zhuo Xu, et~al.
\newblock Rt-trajectory: Robotic task generalization via hindsight trajectory sketches.
\newblock In \emph{ICLR}, 2024.

\bibitem[Jiang et~al.(2025)Jiang, Sun, Huang, Li, Liang, and Xu]{jiang2024robots}
Guangqi Jiang, Yifei Sun, Tao Huang, Huanyu Li, Yongyuan Liang, and Huazhe Xu.
\newblock Robots pre-train robots: Manipulation-centric robotic representation from large-scale robot datasets.
\newblock In \emph{ICLR}, 2025.

\bibitem[Karaev et~al.(2025)Karaev, Makarov, Wang, Neverova, Vedaldi, and Rupprecht]{karaev2024cotracker3simplerbetterpoint}
Nikita Karaev, Yuri Makarov, Jianyuan Wang, Natalia Neverova, Andrea Vedaldi, and Christian Rupprecht.
\newblock Cotracker3: Simpler and better point tracking by pseudo-labelling real videos.
\newblock In \emph{ICCV}, 2025.

\bibitem[Karamcheti et~al.(2024)Karamcheti, Nair, Balakrishna, Liang, Kollar, and Sadigh]{karamcheti2024prismatic}
Siddharth Karamcheti, Suraj Nair, Ashwin Balakrishna, Percy Liang, Thomas Kollar, and Dorsa Sadigh.
\newblock Prismatic vlms: Investigating the design space of visually-conditioned language models.
\newblock In \emph{ICML}, 2024.

\bibitem[Khazatsky et~al.(2024)Khazatsky, Pertsch, Nair, Balakrishna, Dasari, Karamcheti, Nasiriany, Srirama, Chen, Ellis, et~al.]{khazatsky2024droid}
Alexander Khazatsky, Karl Pertsch, Suraj Nair, Ashwin Balakrishna, Sudeep Dasari, Siddharth Karamcheti, Soroush Nasiriany, Mohan~Kumar Srirama, Lawrence~Yunliang Chen, Kirsty Ellis, et~al.
\newblock Droid: A large-scale in-the-wild robot manipulation dataset.
\newblock In \emph{RSS}, 2024.

\bibitem[Kim et~al.(2025{\natexlab{a}})Kim, Kang, Kang, Cho, Kim, and Lee]{kim2025uniskill}
Hanjung Kim, Jaehyun Kang, Hyolim Kang, Meedeum Cho, Seon~Joo Kim, and Youngwoon Lee.
\newblock Uniskill: Imitating human videos via cross-embodiment skill representations.
\newblock In \emph{CoRL}, 2025{\natexlab{a}}.

\bibitem[Kim et~al.(2025{\natexlab{b}})Kim, Pertsch, Karamcheti, Xiao, Balakrishna, Nair, Rafailov, Foster, Sanketi, Vuong, et~al.]{kim2024openvla}
Moo~Jin Kim, Karl Pertsch, Siddharth Karamcheti, Ted Xiao, Ashwin Balakrishna, Suraj Nair, Rafael Rafailov, Ethan~P Foster, Pannag~R Sanketi, Quan Vuong, et~al.
\newblock Openvla: An open-source vision-language-action model.
\newblock In \emph{CoRL}, 2025{\natexlab{b}}.

\bibitem[Ko et~al.(2024)Ko, Mao, Du, Sun, and Tenenbaum]{ko2023learning}
Po-Chen Ko, Jiayuan Mao, Yilun Du, Shao-Hua Sun, and Joshua~B Tenenbaum.
\newblock Learning to act from actionless videos through dense correspondences.
\newblock In \emph{ICLR}, 2024.

\bibitem[Kumar et~al.(2025)Kumar, Huang, Walmer, Rambhatla, and Shrivastava]{kumar2025trokens}
Pulkit Kumar, Shuaiyi Huang, Matthew Walmer, Sai~Saketh Rambhatla, and Abhinav Shrivastava.
\newblock Trokens: Semantic-aware relational trajectory tokens for few-shot action recognition.
\newblock In \emph{ICCV}, 2025.

\bibitem[Lee et~al.(2025)Lee, Duan, Fang, Deng, Liu, Li, Fang, Zhang, Wang, Lee, Han, Pumacay, Wu, Hendrix, Farley, VanderBilt, Farhadi, Fox, and Krishna]{lee2025molmoactactionreasoningmodels}
Jason Lee, Jiafei Duan, Haoquan Fang, Yuquan Deng, Shuo Liu, Boyang Li, Bohan Fang, Jieyu Zhang, Yi~Ru Wang, Sangho Lee, Winson Han, Wilbert Pumacay, Angelica Wu, Rose Hendrix, Karen Farley, Eli VanderBilt, Ali Farhadi, Dieter Fox, and Ranjay Krishna.
\newblock Molmoact: Action reasoning models that can reason in space.
\newblock \emph{arXiv preprint arXiv:2508.07917}, 2025.

\bibitem[Li et~al.(2025{\natexlab{a}})Li, Sun, Hu, Ta, Barry, Konidaris, and Fu]{li2025novaflow}
Hongyu Li, Lingfeng Sun, Yafei Hu, Duy Ta, Jennifer Barry, George Konidaris, and Jiahui Fu.
\newblock Novaflow: Zero-shot manipulation via actionable flow from generated videos.
\newblock \emph{arXiv preprint arXiv:2510.08568}, 2025{\natexlab{a}}.

\bibitem[Li et~al.(2025{\natexlab{b}})Li, Deng, Zhang, Jang, Memmel, Garrett, Ramos, Fox, Li, Gupta, and Goyal]{li2025hamster}
Yi Li, Yuquan Deng, Jesse Zhang, Joel Jang, Marius Memmel, Caelan~Reed Garrett, Fabio Ramos, Dieter Fox, Anqi Li, Abhishek Gupta, and Ankit Goyal.
\newblock Hamster: Hierarchical action models for open-world robot manipulation.
\newblock In \emph{ICLR}, 2025{\natexlab{b}}.

\bibitem[Li et~al.(2025{\natexlab{c}})Li, Tucker, Cole, Wang, Jin, Ye, Kanazawa, Holynski, and Snavely]{li2025megasam}
Zhengqi Li, Richard Tucker, Forrester Cole, Qianqian Wang, Linyi Jin, Vickie Ye, Angjoo Kanazawa, Aleksander Holynski, and Noah Snavely.
\newblock Megasam: Accurate, fast and robust structure and motion from casual dynamic videos.
\newblock In \emph{CVPR}, 2025{\natexlab{c}}.

\bibitem[Liang et~al.(2024)Liang, Xu, Hu, Jiang, Huang, and Xu]{liang2024make}
Yongyuan Liang, Tingqiang Xu, Kaizhe Hu, Guangqi Jiang, Furong Huang, and Huazhe Xu.
\newblock Make-an-agent: A generalizable policy network generator with behavior-prompted diffusion.
\newblock In \emph{NeurIPS}, 2024.

\bibitem[Liu et~al.(2023)Liu, Zhu, Gao, Feng, Liu, Zhu, and Stone]{liu2023libero}
Bo Liu, Yifeng Zhu, Chongkai Gao, Yihao Feng, Qiang Liu, Yuke Zhu, and Peter Stone.
\newblock Libero: Benchmarking knowledge transfer for lifelong robot learning.
\newblock In \emph{NeurIPS}, 2023.

\bibitem[Mandlekar et~al.(2018)Mandlekar, Zhu, Garg, Booher, Spero, Tung, Gao, Emmons, Gupta, Orbay, et~al.]{mandlekar2018roboturk}
Ajay Mandlekar, Yuke Zhu, Animesh Garg, Jonathan Booher, Max Spero, Albert Tung, Julian Gao, John Emmons, Anchit Gupta, Emre Orbay, et~al.
\newblock Roboturk: A crowdsourcing platform for robotic skill learning through imitation.
\newblock In \emph{CoRL}, 2018.

\bibitem[Mandlekar et~al.(2021)Mandlekar, Xu, Wong, Nasiriany, Wang, Kulkarni, Fei-Fei, Savarese, Zhu, and Mart{\'\i}n-Mart{\'\i}n]{mandlekar2021matters}
Ajay Mandlekar, Danfei Xu, Josiah Wong, Soroush Nasiriany, Chen Wang, Rohun Kulkarni, Li Fei-Fei, Silvio Savarese, Yuke Zhu, and Roberto Mart{\'\i}n-Mart{\'\i}n.
\newblock What matters in learning from offline human demonstrations for robot manipulation.
\newblock In \emph{CoRL}, 2021.

\bibitem[Niu et~al.(2025)Niu, Sharma, Xue, Biamby, Zhang, Ji, Darrell, and Herzig]{niu2025pre}
Dantong Niu, Yuvan Sharma, Haoru Xue, Giscard Biamby, Junyi Zhang, Ziteng Ji, Trevor Darrell, and Roei Herzig.
\newblock Pre-training auto-regressive robotic models with 4d representations.
\newblock In \emph{ICML}, 2025.

\bibitem[O’Neill et~al.(2024)O’Neill, Rehman, Maddukuri, Gupta, Padalkar, Lee, Pooley, Gupta, Mandlekar, Jain, et~al.]{o2024open}
Abby O’Neill, Abdul Rehman, Abhiram Maddukuri, Abhishek Gupta, Abhishek Padalkar, Abraham Lee, Acorn Pooley, Agrim Gupta, Ajay Mandlekar, Ajinkya Jain, et~al.
\newblock Open x-embodiment: Robotic learning datasets and rt-x models: Open x-embodiment collaboration.
\newblock In \emph{ICRA}, 2024.

\bibitem[Qu et~al.(2025)Qu, Song, Chen, Chen, Gao, Ye, Lv, Shi, Ren, Ruan, et~al.]{qu2025embodiedonevision}
Delin Qu, Haoming Song, Qizhi Chen, Zhaoqing Chen, Xianqiang Gao, Xinyi Ye, Qi Lv, Modi Shi, Guanghui Ren, Cheng Ruan, et~al.
\newblock Embodiedonevision: Interleaved vision-text-action pretraining for general robot control.
\newblock \emph{arXiv preprint arXiv:2508.21112}, 2025.

\bibitem[Raffel et~al.(2020)Raffel, Shazeer, Roberts, Lee, Narang, Matena, Zhou, Li, and Liu]{raffel2020exploring}
Colin Raffel, Noam Shazeer, Adam Roberts, Katherine Lee, Sharan Narang, Michael Matena, Yanqi Zhou, Wei Li, and Peter~J Liu.
\newblock Exploring the limits of transfer learning with a unified text-to-text transformer.
\newblock \emph{Journal of machine learning research}, 21\penalty0 (140):\penalty0 1--67, 2020.

\bibitem[Ranasinghe et~al.(2025)Ranasinghe, Li, Nguyen, Mata, Park, and Ryoo]{ranasinghe2025langtomo}
Kanchana Ranasinghe, Xiang Li, E-Ro Nguyen, Cristina Mata, Jongwoo Park, and Michael~S Ryoo.
\newblock Pixel motion as universal representation for robot control.
\newblock \emph{arXiv preprint arXiv:2505.07817}, 2025.

\bibitem[Seita et~al.(2023)Seita, Wang, Shetty, Li, Erickson, and Held]{seita2023toolflownet}
Daniel Seita, Yufei Wang, Sarthak~J Shetty, Edward~Yao Li, Zackory Erickson, and David Held.
\newblock Toolflownet: Robotic manipulation with tools via predicting tool flow from point clouds.
\newblock In \emph{CoRL}, 2023.

\bibitem[Sim{\'e}oni et~al.(2025)Sim{\'e}oni, Vo, Seitzer, Baldassarre, Oquab, Jose, Khalidov, Szafraniec, Yi, Ramamonjisoa, et~al.]{simeoni2025dinov3}
Oriane Sim{\'e}oni, Huy~V Vo, Maximilian Seitzer, Federico Baldassarre, Maxime Oquab, Cijo Jose, Vasil Khalidov, Marc Szafraniec, Seungeun Yi, Micha{\"e}l Ramamonjisoa, et~al.
\newblock Dinov3.
\newblock \emph{arXiv preprint arXiv:2508.10104}, 2025.

\bibitem[Sun et~al.(2023)Sun, Ma, Madaan, Bonatti, Huang, and Kapoor]{sun2023smart}
Yanchao Sun, Shuang Ma, Ratnesh Madaan, Rogerio Bonatti, Furong Huang, and Ashish Kapoor.
\newblock Smart: Self-supervised multi-task pretraining with control transformers.
\newblock In \emph{ICLR}, 2023.

\bibitem[Vecerik et~al.(2024)Vecerik, Doersch, Yang, Davchev, Aytar, Zhou, Hadsell, Agapito, and Scholz]{vecerik2024robotap}
Mel Vecerik, Carl Doersch, Yi Yang, Todor Davchev, Yusuf Aytar, Guangyao Zhou, Raia Hadsell, Lourdes Agapito, and Jon Scholz.
\newblock Robotap: Tracking arbitrary points for few-shot visual imitation.
\newblock In \emph{ICRA}, 2024.

\bibitem[Walke et~al.(2023)Walke, Black, Zhao, Vuong, Zheng, Hansen-Estruch, He, Myers, Kim, Du, et~al.]{walke2023bridgedata}
Homer~Rich Walke, Kevin Black, Tony~Z Zhao, Quan Vuong, Chongyi Zheng, Philippe Hansen-Estruch, Andre~Wang He, Vivek Myers, Moo~Jin Kim, Max Du, et~al.
\newblock Bridgedata v2: A dataset for robot learning at scale.
\newblock In \emph{CoRL}, 2023.

\bibitem[Wang et~al.(2025)Wang, Chen, Karaev, Vedaldi, Rupprecht, and Novotny]{vggt}
Jianyuan Wang, Minghao Chen, Nikita Karaev, Andrea Vedaldi, Christian Rupprecht, and David Novotny.
\newblock Vggt: Visual geometry grounded transformer.
\newblock In \emph{CVPR}, 2025.

\bibitem[Wen et~al.(2024)Wen, Lin, So, Chen, Dou, Gao, and Abbeel]{wen2023any}
Chuan Wen, Xingyu Lin, John So, Kai Chen, Qi Dou, Yang Gao, and Pieter Abbeel.
\newblock Any-point trajectory modeling for policy learning.
\newblock In \emph{RSS}, 2024.

\bibitem[Xiao et~al.(2025)Xiao, Wang, Xue, Karaev, Makarov, Kang, Zhu, Bao, Shen, and Zhou]{xiao2025spatialtrackerv2}
Yuxi Xiao, Jianyuan Wang, Nan Xue, Nikita Karaev, Yuri Makarov, Bingyi Kang, Xing Zhu, Hujun Bao, Yujun Shen, and Xiaowei Zhou.
\newblock Spatialtrackerv2: 3d point tracking made easy.
\newblock In \emph{ICCV}, 2025.

\bibitem[Xu et~al.(2024)Xu, Xu, Xu, Chi, Wetzstein, Veloso, and Song]{xu2024flowcrossdomainmanipulationinterface}
Mengda Xu, Zhenjia Xu, Yinghao Xu, Cheng Chi, Gordon Wetzstein, Manuela Veloso, and Shuran Song.
\newblock Flow as the cross-domain manipulation interface.
\newblock In \emph{CoRL}, 2024.

\bibitem[Yang et~al.(2025{\natexlab{a}})Yang, Tan, Wu, Zheng, Peng, Liang, Gu, Cai, Ye, Jang, et~al.]{yang2025magma}
Jianwei Yang, Reuben Tan, Qianhui Wu, Ruijie Zheng, Baolin Peng, Yongyuan Liang, Yu Gu, Mu Cai, Seonghyeon Ye, Joel Jang, et~al.
\newblock Magma: A foundation model for multimodal ai agents.
\newblock In \emph{CVPR}, 2025{\natexlab{a}}.

\bibitem[Yang et~al.(2025{\natexlab{b}})Yang, Teng, Zheng, Ding, Huang, Xu, Yang, Hong, Zhang, Feng, et~al.]{yang2024cogvideox}
Zhuoyi Yang, Jiayan Teng, Wendi Zheng, Ming Ding, Shiyu Huang, Jiazheng Xu, Yuanming Yang, Wenyi Hong, Xiaohan Zhang, Guanyu Feng, et~al.
\newblock Cogvideox: Text-to-video diffusion models with an expert transformer.
\newblock In \emph{ICLR}, 2025{\natexlab{b}}.

\bibitem[Young et~al.(2021)Young, Gandhi, Tulsiani, Gupta, Abbeel, and Pinto]{young2021visual}
Sarah Young, Dhiraj Gandhi, Shubham Tulsiani, Abhinav Gupta, Pieter Abbeel, and Lerrel Pinto.
\newblock Visual imitation made easy.
\newblock In \emph{CoRL}, 2021.

\bibitem[Yu et~al.(2025{\natexlab{a}})Yu, Zhang, Soora, Huang, Huang, Tokekar, and Gao]{yu2025genflowrl}
Kelin Yu, Sheng Zhang, Harshit Soora, Furong Huang, Heng Huang, Pratap Tokekar, and Ruohan Gao.
\newblock Genflowrl: Shaping rewards with generative object-centric flow in visual reinforcement learning.
\newblock In \emph{ICCV}, 2025{\natexlab{a}}.

\bibitem[Yu et~al.(2025{\natexlab{b}})Yu, Bhaskar, Singh, Mahammad, and Tokekar]{yu2025sketch}
Peihong Yu, Amisha Bhaskar, Anukriti Singh, Zahiruddin Mahammad, and Pratap Tokekar.
\newblock Sketch-to-skill: Bootstrapping robot learning with human drawn trajectory sketches.
\newblock In \emph{RSS}, 2025{\natexlab{b}}.

\bibitem[Yuan et~al.(2024)Yuan, Duan, Blukis, Pumacay, Krishna, Murali, Mousavian, and Fox]{yuan2024robopoint}
Wentao Yuan, Jiafei Duan, Valts Blukis, Wilbert Pumacay, Ranjay Krishna, Adithyavairavan Murali, Arsalan Mousavian, and Dieter Fox.
\newblock Robopoint: A vision-language model for spatial affordance prediction for robotics.
\newblock In \emph{CoRL}, 2024.

\bibitem[Yuan et~al.(2025{\natexlab{a}})Yuan, Cui, Chen, Dong, Ni, Kou, Liu, Li, Zheng, and Hao]{yuan2025seeing}
Yifu Yuan, Haiqin Cui, Yibin Chen, Zibin Dong, Fei Ni, Longxin Kou, Jinyi Liu, Pengyi Li, Yan Zheng, and Jianye Hao.
\newblock From seeing to doing: Bridging reasoning and decision for robotic manipulation.
\newblock \emph{arXiv preprint arXiv:2505.08548}, 2025{\natexlab{a}}.

\bibitem[Yuan et~al.(2025{\natexlab{b}})Yuan, Cui, Huang, Chen, Ni, Dong, Li, Zheng, and Hao]{yuan2025embodied}
Yifu Yuan, Haiqin Cui, Yaoting Huang, Yibin Chen, Fei Ni, Zibin Dong, Pengyi Li, Yan Zheng, and Jianye Hao.
\newblock Embodied-r1: Reinforced embodied reasoning for general robotic manipulation.
\newblock \emph{arXiv preprint arXiv:2508.13998}, 2025{\natexlab{b}}.

\bibitem[Zhai et~al.(2023)Zhai, Mustafa, Kolesnikov, and Beyer]{zhai2023sigmoid}
Xiaohua Zhai, Basil Mustafa, Alexander Kolesnikov, and Lucas Beyer.
\newblock Sigmoid loss for language image pre-training.
\newblock In \emph{ICCV}, 2023.

\bibitem[Zhang et~al.(2025)Zhang, Ke, Harley, and Fragkiadaki]{zhang2025tapip3dtrackingpointpersistent}
Bowei Zhang, Lei Ke, Adam~W. Harley, and Katerina Fragkiadaki.
\newblock Tapip3d: Tracking any point in persistent 3d geometry.
\newblock In \emph{NeurIPS}, 2025.

\bibitem[Zhen et~al.(2025)Zhen, Sun, Zhang, Li, Zhou, Du, and Gan]{zhen2025learning}
Haoyu Zhen, Qiao Sun, Hongxin Zhang, Junyan Li, Siyuan Zhou, Yilun Du, and Chuang Gan.
\newblock Learning 4d embodied world models.
\newblock In \emph{ICCV}, 2025.

\bibitem[Zheng et~al.(2023)Zheng, Wang, Sun, Ma, Zhao, Xu, Daum{\'e}~III, and Huang]{zheng2023texttt}
Ruijie Zheng, Xiyao Wang, Yanchao Sun, Shuang Ma, Jieyu Zhao, Huazhe Xu, Hal Daum{\'e}~III, and Furong Huang.
\newblock Taco: Temporal latent action-driven contrastive loss for visual reinforcement learning.
\newblock In \emph{NeurIPS}, 2023.

\bibitem[Zheng et~al.(2024)Zheng, Liang, Wang, Ma, Daum{\'e}~III, Xu, Langford, Palanisamy, Basu, and Huang]{zheng2024premier}
Ruijie Zheng, Yongyuan Liang, Xiyao Wang, Shuang Ma, Hal Daum{\'e}~III, Huazhe Xu, John Langford, Praveen Palanisamy, Kalyan~Shankar Basu, and Furong Huang.
\newblock Premier-taco is a few-shot policy learner: Pretraining multitask representation via temporal action-driven contrastive loss.
\newblock In \emph{ICML}, 2024.

\bibitem[Zheng et~al.(2025)Zheng, Liang, Huang, Gao, Daum{\'e}~III, Kolobov, Huang, and Yang]{zheng2024tracevla}
Ruijie Zheng, Yongyuan Liang, Shuaiyi Huang, Jianfeng Gao, Hal Daum{\'e}~III, Andrey Kolobov, Furong Huang, and Jianwei Yang.
\newblock Tracevla: Visual trace prompting enhances spatial-temporal awareness for generalist robotic policies.
\newblock In \emph{ICLR}, 2025.

\bibitem[Zhi et~al.(2025)Zhi, Chen, Zhou, Dong, Wu, Han, and Tan]{zhi20253dflowaction}
Hongyan Zhi, Peihao Chen, Siyuan Zhou, Yubo Dong, Quanxi Wu, Lei Han, and Mingkui Tan.
\newblock 3dflowaction: Learning cross-embodiment manipulation from 3d flow world model.
\newblock \emph{arXiv preprint arXiv:2506.06199}, 2025.

\bibitem[Zhou et~al.(2025)Zhou, Pan, LeCun, and Pinto]{Zhou2025Dino-wm}
Gaoyue Zhou, Hengkai Pan, Yann LeCun, and Lerrel Pinto.
\newblock Dino-wm: World models on pre-trained visual features enable zero-shot planning.
\newblock In \emph{ICML}, 2025.

\bibitem[Zitkovich et~al.(2023)Zitkovich, Yu, Xu, Xu, Xiao, Xia, Wu, Wohlhart, Welker, Wahid, et~al.]{zitkovich2023rt}
Brianna Zitkovich, Tianhe Yu, Sichun Xu, Peng Xu, Ted Xiao, Fei Xia, Jialin Wu, Paul Wohlhart, Stefan Welker, Ayzaan Wahid, et~al.
\newblock Rt-2: Vision-language-action models transfer web knowledge to robotic control.
\newblock In \emph{CoRL}, 2023.

\end{thebibliography}
}

\clearpage
\section*{Appendix}
\appendix

\section{Prompts for Task Instruction Generation}

To obtain consistent and diverse task instructions from video segments in \traceforgelogo TraceForge, we use a vision-language model (VLM) to transform representative video frames into three complementary instruction styles. As described in the main manuscript, each event chunk is paired with (i) a concise imperative command, (ii) a stepwise manipulation instruction, and (iii) a natural, human-like request. When human-written instructions are already available, we preserve them and augment them with these additional variants. Otherwise, we sample frames from the beginning, middle, and end of the chunk and prompt the VLM to generate instructions following the structured specification below.

\begin{tcolorbox}[
    breakable,
    colback=orange!5,
    colframe=orange!10!,
    coltitle=black,
    sharp corners,
    enhanced,
    boxrule=0.6pt,
    title={Sample Prompt for Instruction Generation},
    fonttitle=\bfseries,
]
You are an expert image analyzer. You will receive a sequence of frames
from a single video episode that records a simple manipulation task.
The frames are ordered chronologically from initial state to final state.

\textbf{GOAL}  
Infer the most likely task being performed in this episode and return the
OUTPUT FORMAT below.

\textbf{TASK INSTRUCTIONS}
\begin{itemize}
    \item IMPORTANT: Do \textbf{NOT} generate a descriptive sentence like
    “The agent is trying to grab the sink faucet.”
    \item Instead, generate instructions as if a human is directly
    commanding a robot or agent to make this situation happen.
    \item Provide exactly three instructions, one for each category:
    \begin{itemize}
        \item \texttt{instruction\_1}: Direct, simple imperative command  
        e.g., “Turn on the faucet.”
        \item \texttt{instruction\_2}: Step-by-step explicit manipulation  
        e.g., “Move your gripper toward the faucet handle.”
        \item \texttt{instruction\_3}: Natural human-like request  
        e.g., “Can you turn on the sink for me?”
    \end{itemize}
    \item All instructions must explicitly state \textbf{what} object is
    manipulated and \textbf{how}.
\end{itemize}

\textbf{Instruction length rules}
\begin{itemize}
    \item \texttt{instruction\_1}: $\leq$ 10 words  
    \item \texttt{instruction\_2}: $\leq$ 20 words  
    \item \texttt{instruction\_3}: $\leq$ 15 words  
\end{itemize}

The operator can be a human, robot, or tool.

\textbf{OUTPUT (JSON ONLY)}
\begin{verbatim}
{
  "instruction_1": 
      "<=10 words 
      imperative command>",
  "instruction_2": 
      "<=20 words 
      stepwise/explicit command>",
  "instruction_3": 
      "<=15 words 
      natural human-like request>"
}
\end{verbatim}

\textbf{POLICIES}
\begin{itemize}
    \item Do NOT invent objects that are not visible.
    \item Prefer specific names if clear (banana), else generic ones
          (container).
    \item All strings must be in English.
    \item Return only valid JSON—no markdown or extra text.
\end{itemize}
\end{tcolorbox}

\section{3D Trace Extraction Accuracy of TraceForge}
\label{traceforge_result}

As discussed in the main text, \traceforgelogo TraceForge provides the
large-scale 3D motion supervision used to train TraceGen. To ensure that this
supervision is reliable, we report a quantitative sanity check of TraceForge's
3D trace extraction accuracy.

We evaluate whether TraceForge recovers 3D trajectories that faithfully reflect
real robot motion by comparing its predicted traces with ground-truth
end-effector trajectories obtained via forward kinematics across nine
teleoperated episodes. Since TraceForge represents motion as a $20\times20$
grid of point trajectories, we identify, for each episode, the predicted point
whose 2D projection is closest to the end-effector in the first frame and treat
its 3D path as the corresponding predicted trace.

Across episodes (13–24.5 seconds each, with an average displacement of
70.96\,cm), TraceForge achieves sub–2.3\,cm endpoint error on all axes
(Table~\ref{tabele:mse_tracegen_error}). These results indicate that the
TraceForge extraction pipeline produces centimeter-level motion accuracy,
providing reliable supervision for training TraceGen.

\begin{table}[h]
\centering
\small
\caption{Absolute endpoint error along the $x,y,z$ axes in camera coordinate between predicted and ground-truth trajectories.}
\label{tabele:mse_tracegen_error}
\begin{tabular}{lccc}
\toprule
Error (cm) &  x & y & z \\
\midrule
{Mean}  & 1.66 & 1.79 & 2.26 \\
{Std}   & 0.82 & 1.82 & 2.69 \\
\bottomrule
\end{tabular}
\end{table}

\section{Model Training Details}
\label{sec:training_details}

This section provides comprehensive details on TraceGen's training configuration, complementing the architecture overview in the main manuscript. TraceGen employs a multi-encoder architecture with DINOv2 and SigLIP for RGB feature extraction, a depth encoder with a learnable stem adapter for geometric information, and T5 for text encoding.

\subsection{Encoder Freezing Strategy}
\label{sec:encoder_freezing_detailed}

To leverage pretrained representations efficiently, we keep all encoders (DINOv2, SigLIP, T5) frozen throughout training and train only the fusion layer and decoder. This design choice follows Prismatic VLM~\cite{karamcheti2024prismatic}, which demonstrated that finetuning visual backbones significantly degrades performance on vision-language tasks. Their analysis revealed that updating pretrained vision encoders during task-specific finetuning leads to catastrophic forgetting of the rich visual priors learned during large-scale pretraining.

By maintaining frozen encoders, TraceGen retains:
\begin{itemize}[leftmargin=*,noitemsep,topsep=0pt]
    \item \textbf{Visual features from DINOv2}, trained via self-supervised learning on large-scale natural images, providing strong geometric and semantic understanding
    \item \textbf{Vision-language alignment from SigLIP}, enabling effective conditioning on text instructions
    \item \textbf{Linguistic representations from T5}, capturing task semantics and manipulation goals
\end{itemize}

The trainable fusion layer and flow-based decoder learn to combine and map these frozen representations to the 3D trace prediction task. This approach substantially reduces the number of trainable parameters, accelerates training, and improves generalization to unseen manipulation scenarios.

\section{Evaluations}
\subsection{Evaluation setup}
We evaluate whether 3D traces predicted by TraceGen enable effective robot manipulation. Experiments are conducted on a Franka Research 3 robot across four tasks:
\begin{itemize}
    \item folding a garment (Clothes)
    \item inserting a tennis ball into a box (Ball)
    \item sweeping trash into a dustpan with a brush (Brush)
    \item placing a LEGO block in the purple region (Block)
\end{itemize}

Given a single RGB-D observation and a text instruction, TraceGen predicts trajectories using 100-step ODE integration with classifier-free guidance (guidance scale 2) for \textit{Brush} and \textit{Clothes}, and without guidance for \textit{Block} and \textit{Ball}. The predicted trajectories are converted into joint-space commands through inverse kinematics after transforming them from the camera frame to the robot base frame \(T_{\mathrm{base}\leftarrow \mathrm{cam}_{\mathrm{ref}}}\).
For all methods, the predicted $z$ values are rescaled to match the measured depth maps.

\subsection{Depth rescaling}

To align the predicted trace depths with the original sensor depth, we apply a depth-rescaling procedure. Similar to NovaFlow, which computes a single scaling factor based on the median depth of the initial ground-truth map, we also begin by estimating a global scaling factor between the predicted depth and the sensor depth.

However, unlike the environments used in NovaFlow, our settings exhibit much larger variations and more frequent movements along the depth axis. We observed that using a single median-based scalar often leads to substantial depth estimation errors in such scenarios.

To address this, instead of relying on a single global statistic, we compute a pixel-wise depth rescaling map by directly comparing the predicted and sensor depth maps across all pixels. We then apply a Gaussian blur to this map to obtain a smooth depth-rescaling field, and multiply this smoothed map with the predicted 3D trace to correct its z-values.

\subsection{Implementation details for the baselines}

\paragraph{3DFlowaction}
The official 3DFlowAction implementation relies on a filtering pipeline to extract object masks. This pipeline combines a language-conditioned object detector with a heuristic process designed to remove the robot gripper. However, when only a single image is provided as input, the detector often fails to identify the target object reliably or to filter out the robot end-effector. As a result, the predicted masks frequently include parts of the robot itself.

To mitigate this issue, and consistent with the official implementation, we provide a short sequence of images containing minimal robot motion. These slight temporal changes help the filtering pipeline correctly identify the robot gripper, allowing the final filtered region to closely match the ground-truth object mask. To ensure the generated object mask aligned with the expected robot movement, we manually checked whether the detected bounding box is aligned with the ground truth bounding box of the target objects.

\begin{figure}[t]
    \centering

    \begin{subfigure}[t]{0.48\textwidth}
        \centering
        \includegraphics[width=\textwidth]{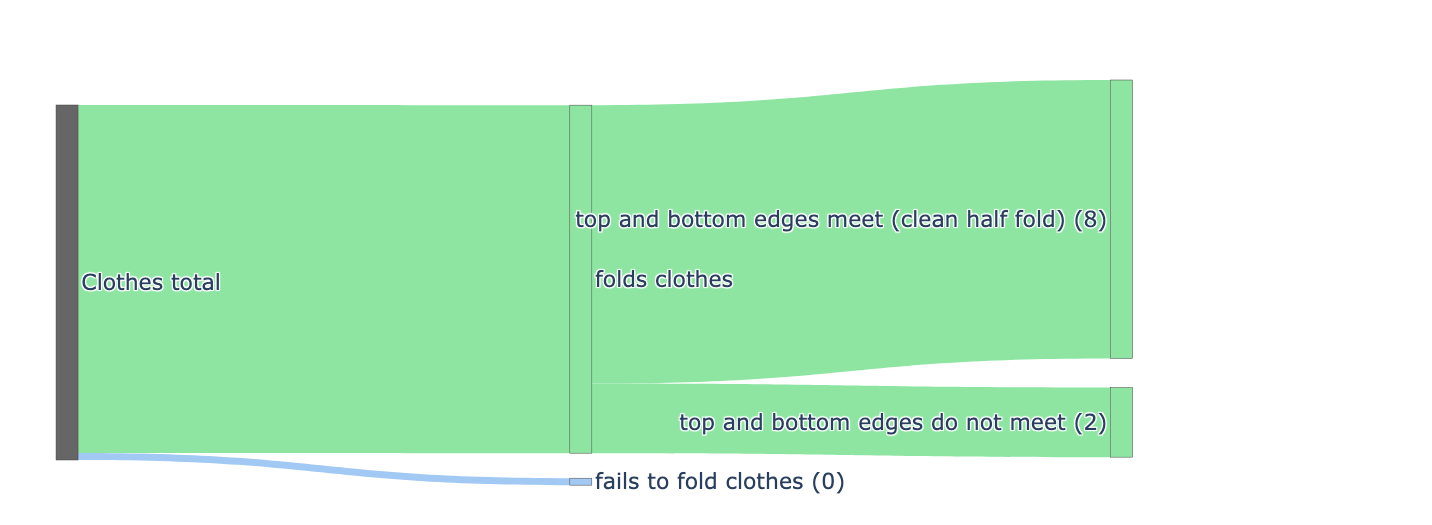}
        \caption{TraceGen (Warmed up with robot demo)}
    \end{subfigure}%
    \par\smallskip
    \begin{subfigure}[t]{0.48\textwidth}
        \centering
        \includegraphics[width=\textwidth]{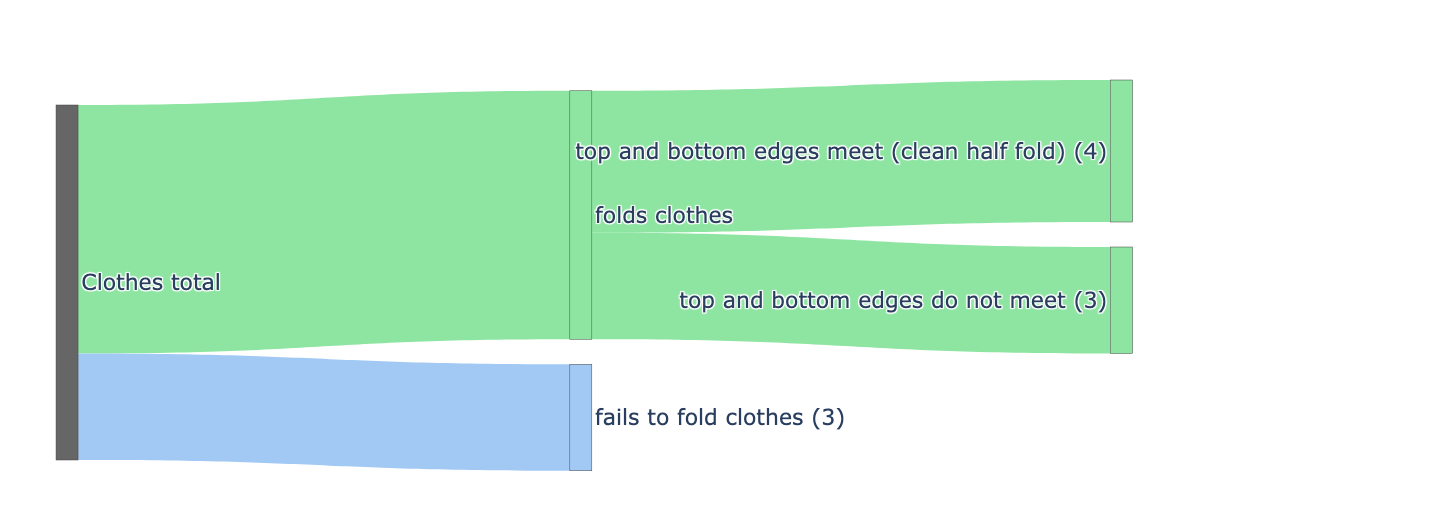}
        \caption{TraceGen (Warmed up with human demo)}
    \end{subfigure}

    \par\smallskip
    \begin{subfigure}[t]{0.48\textwidth}
        \centering
        \includegraphics[width=\textwidth]{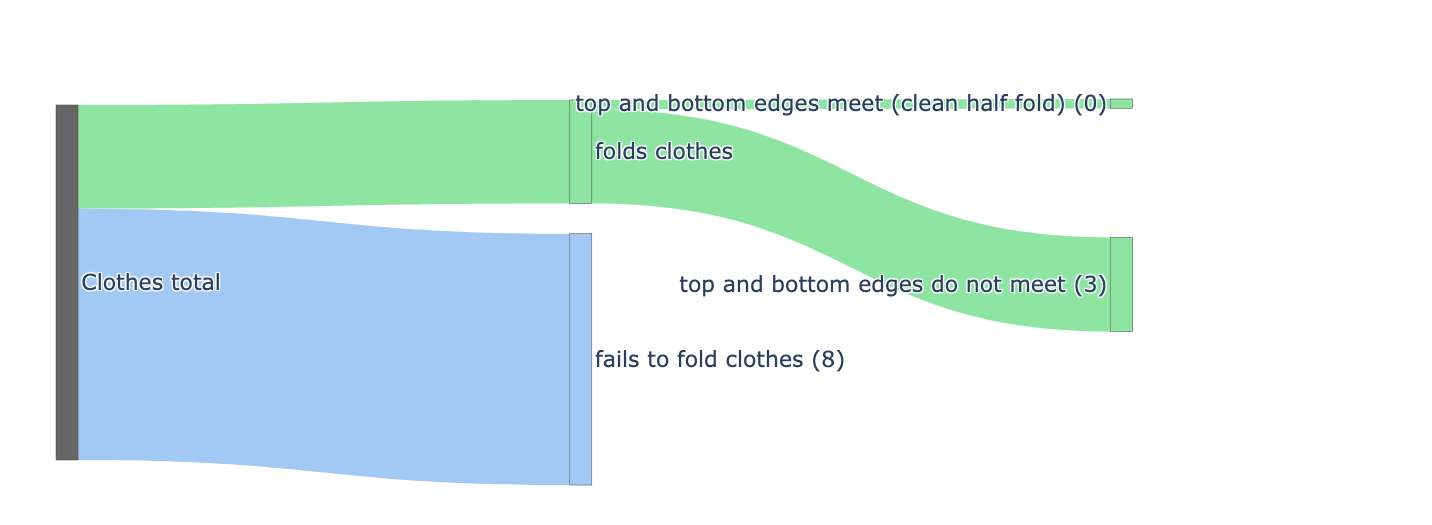}
        \caption{NovaFlow (Veo3.1)}
    \end{subfigure}%
    \par\smallskip
    \begin{subfigure}[t]{0.48\textwidth}
        \centering
        \includegraphics[width=\textwidth]{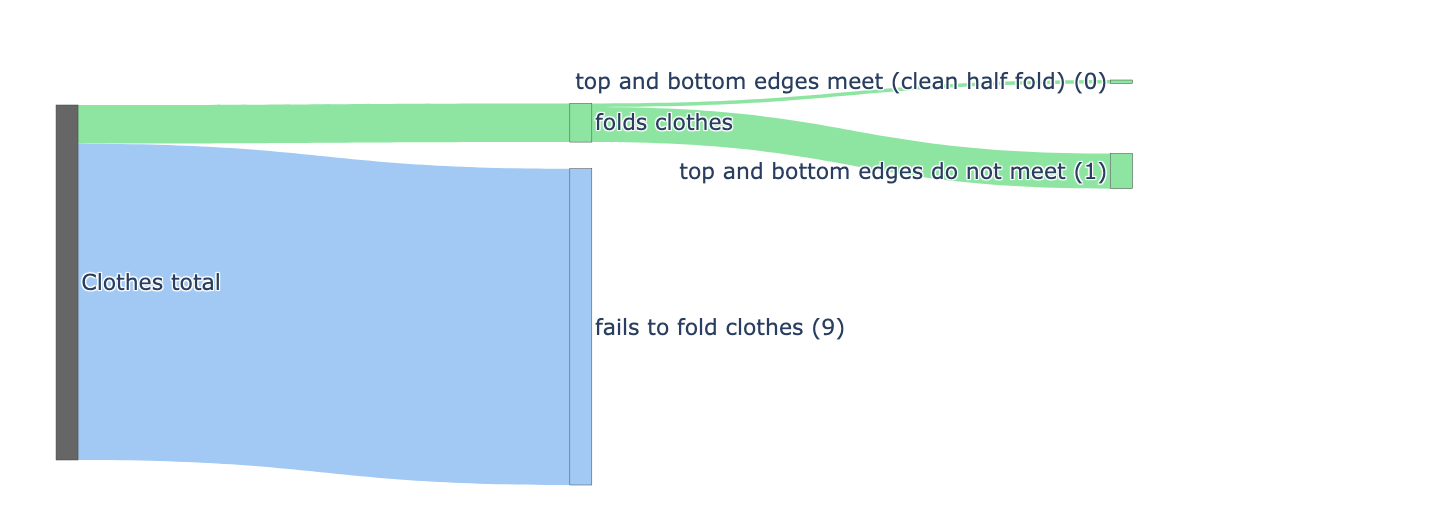}
        \caption{NovaFlow (Wan2.2)}
    \end{subfigure}

    \caption{Failure-mode breakdown for the \emph{Clothes} task.}
    \label{fig:clothes_failure_modes}
\end{figure}

\begin{figure}[t]
    \centering

    \begin{subfigure}[t]{0.48\textwidth}
        \centering
        \includegraphics[width=\textwidth]{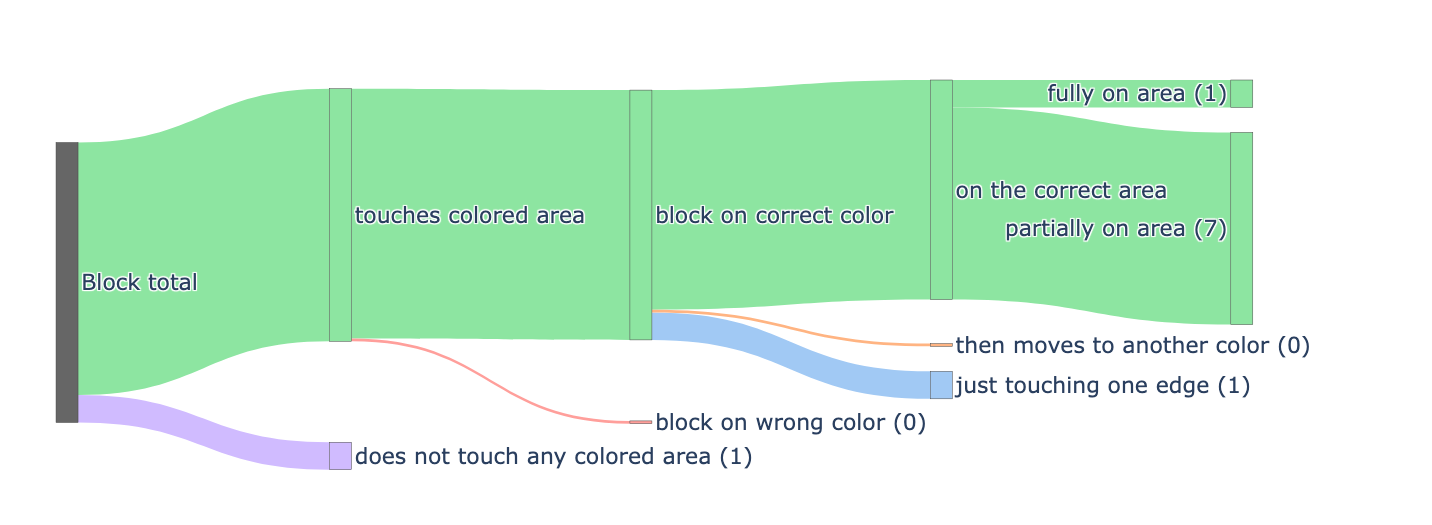}
        \caption{TraceGen (Warmed up with robot demo)}
    \end{subfigure}%
    \par\smallskip
    \begin{subfigure}[t]{0.48\textwidth}
        \centering
        \includegraphics[width=\textwidth]{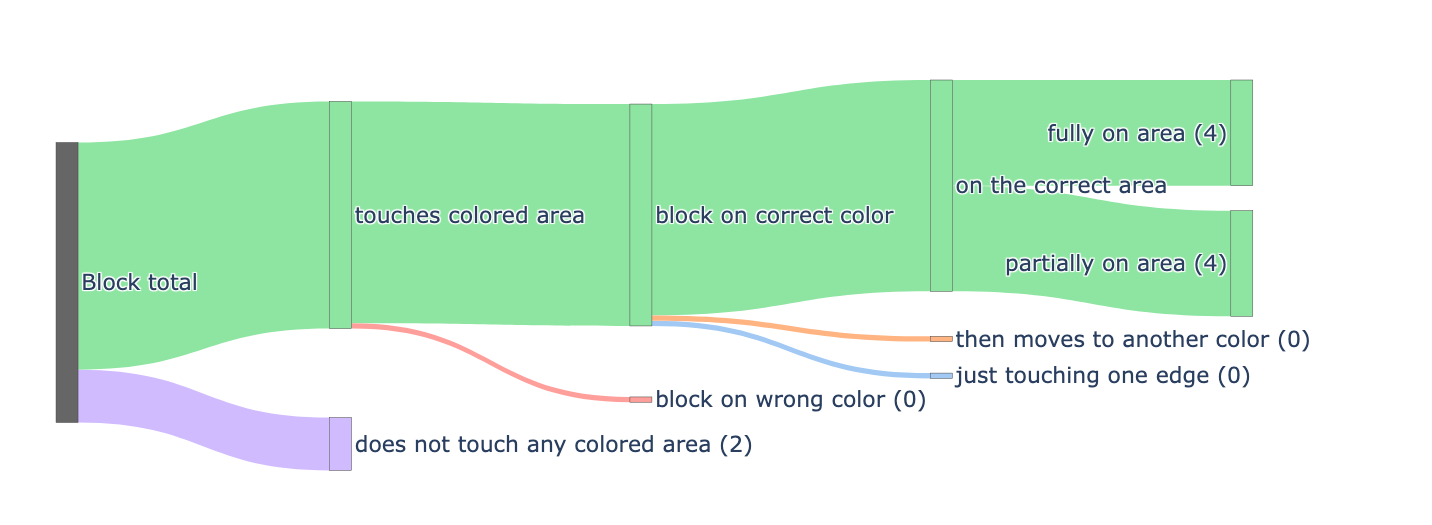}
        \caption{TraceGen (Warmed up with human demo)}
    \end{subfigure}

    \par\smallskip
    \begin{subfigure}[t]{0.48\textwidth}
        \centering
        \includegraphics[width=\textwidth]{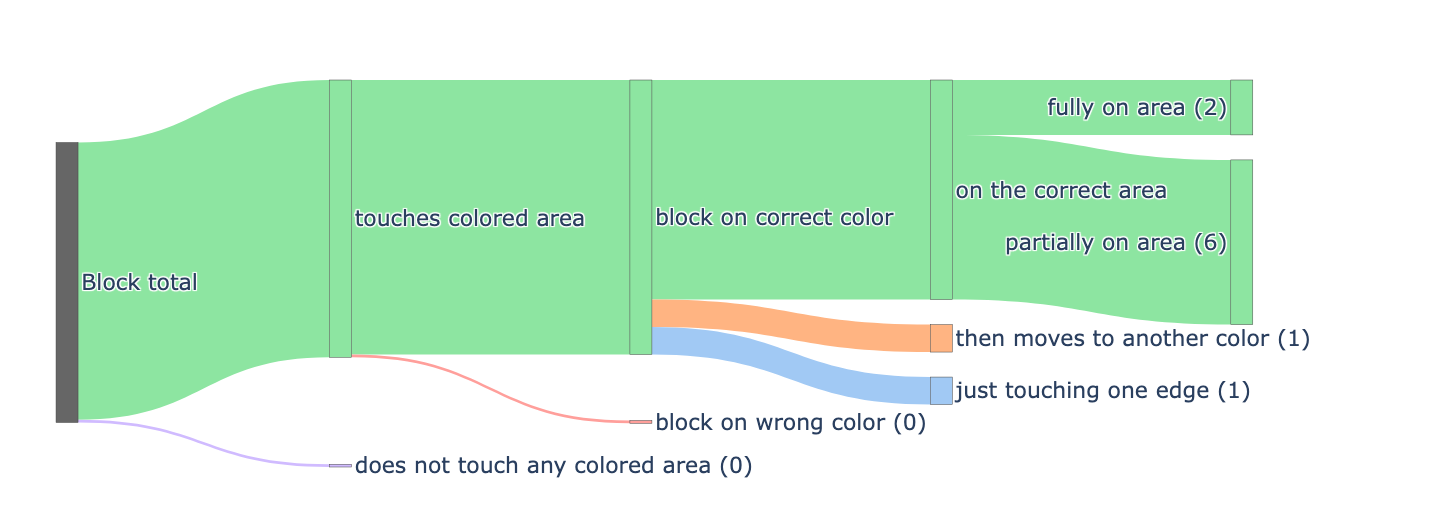}
        \caption{NovaFlow (Veo3.1)}
    \end{subfigure}%
    \par\smallskip
    \begin{subfigure}[t]{0.48\textwidth}
        \centering
        \includegraphics[width=\textwidth]{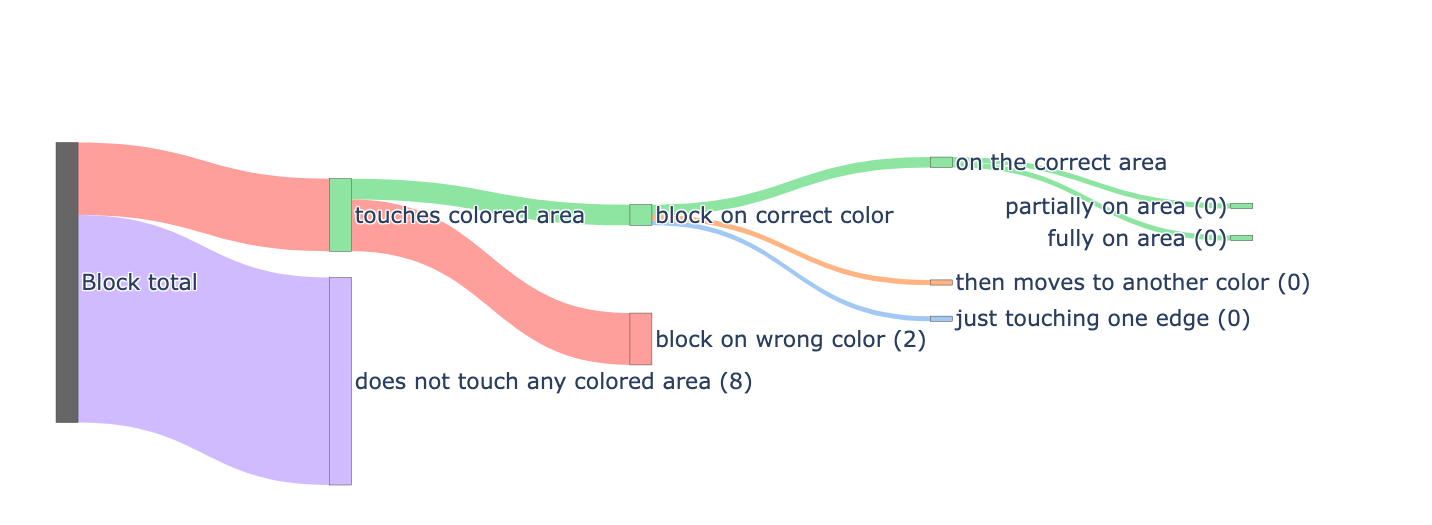}
        \caption{NovaFlow (Wan2.2)}
    \end{subfigure}

    \caption{Failure-mode breakdown for the \emph{Block} task.}
    \label{fig:block_failure_modes}
\end{figure}

\begin{figure}[t]
    \centering

    \begin{subfigure}[t]{0.48\textwidth}
        \centering
        \includegraphics[width=\textwidth]{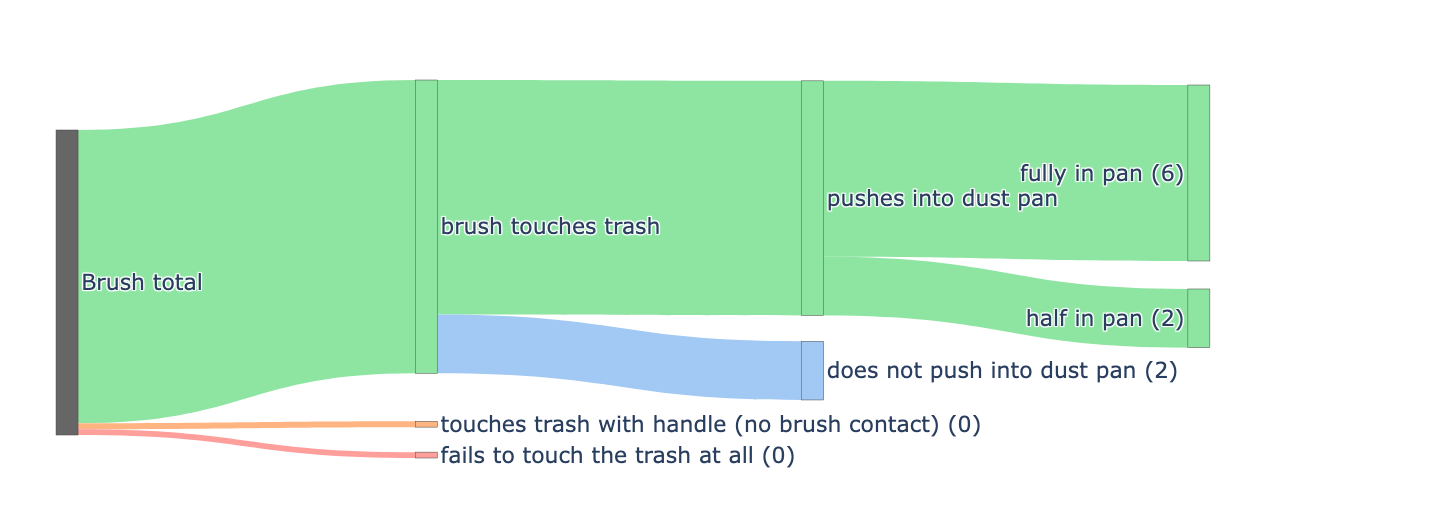}
        \caption{TraceGen (Warmed up with robot demo)}
    \end{subfigure}%
    \par\smallskip
    \begin{subfigure}[t]{0.48\textwidth}
        \centering
        \includegraphics[width=\textwidth]{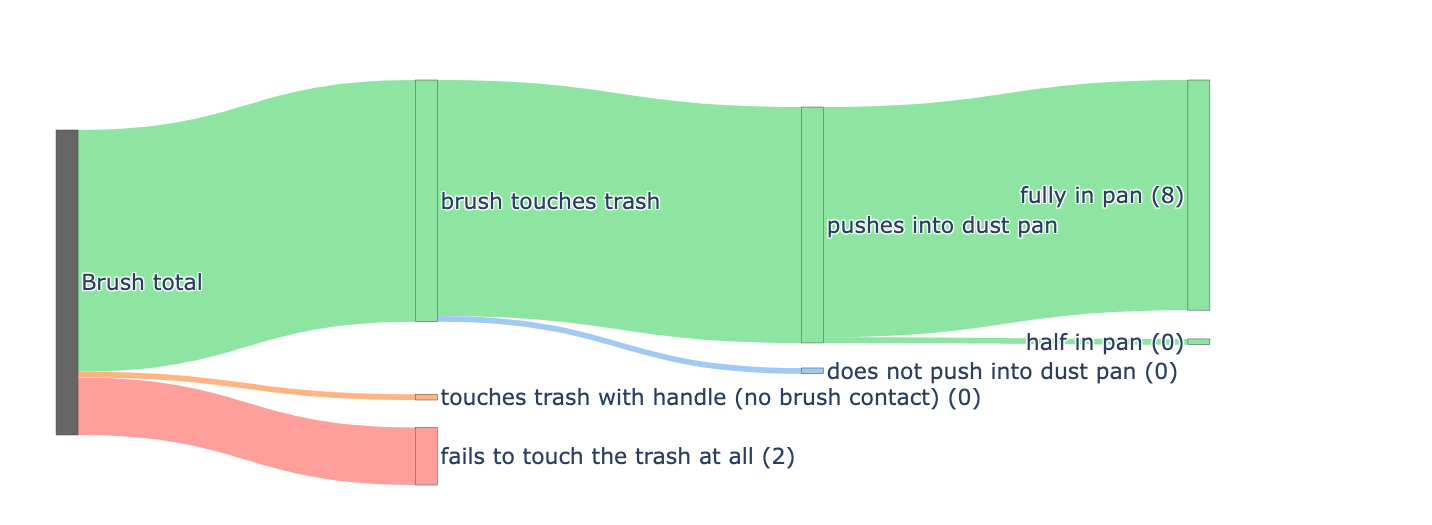}
        \caption{TraceGen (Warmed up with human demo)}
    \end{subfigure}

    \par\smallskip
    \begin{subfigure}[t]{0.48\textwidth}
        \centering
        \includegraphics[width=\textwidth]{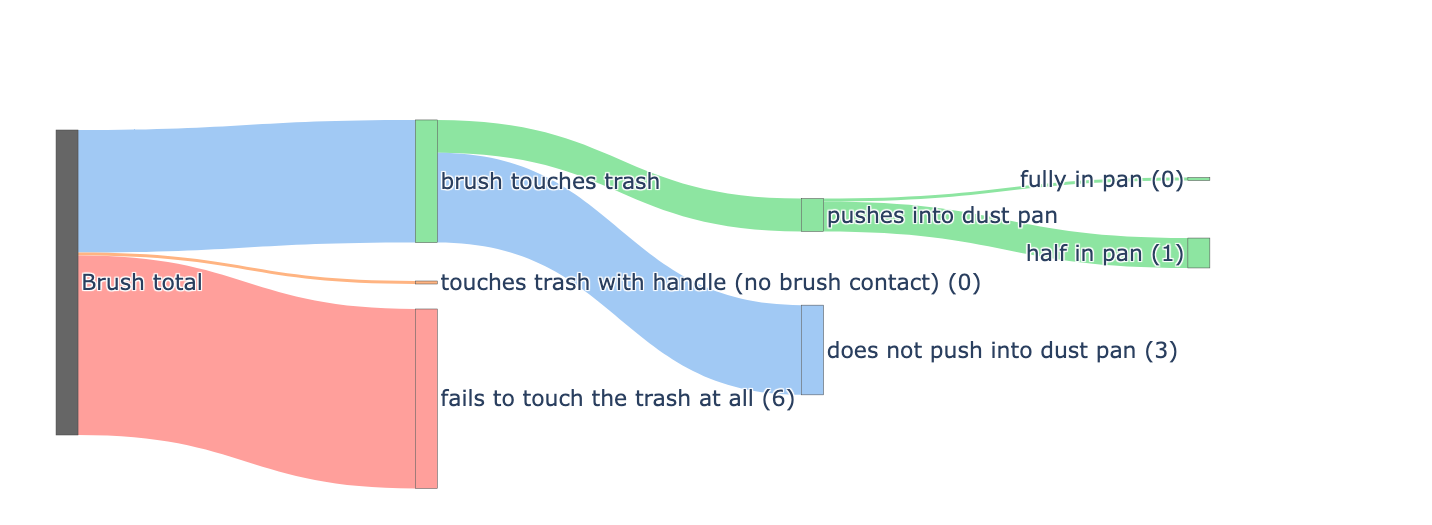}
        \caption{NovaFlow (Veo3.1)}
    \end{subfigure}%
    \par\smallskip
    \begin{subfigure}[t]{0.48\textwidth}
        \centering
        \includegraphics[width=\textwidth]{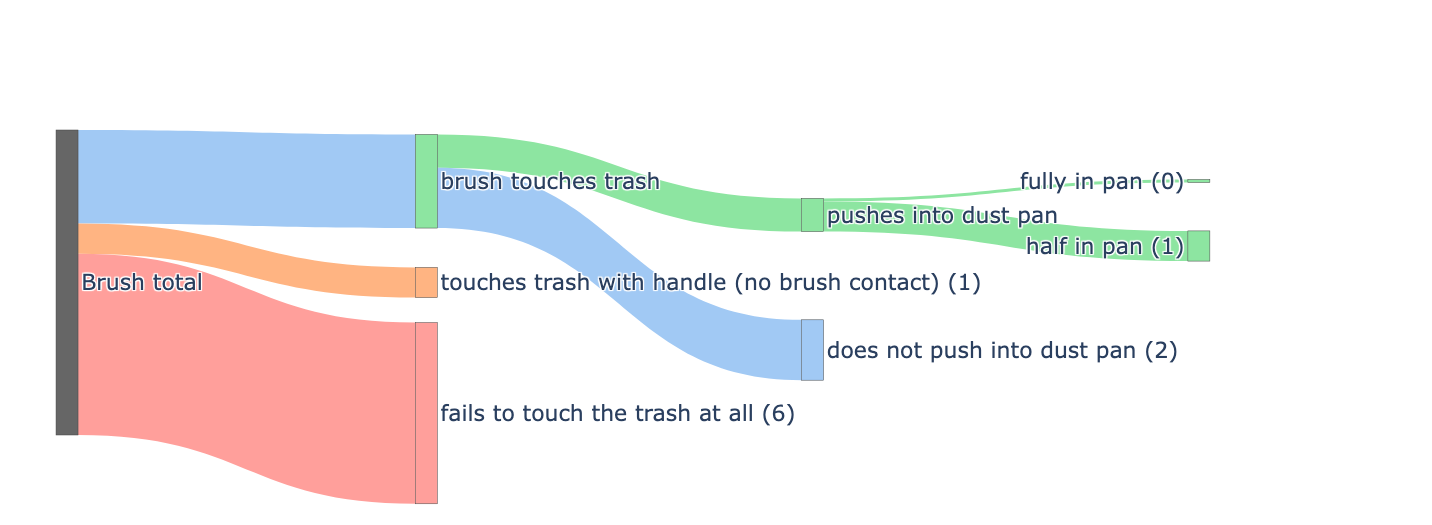}
        \caption{NovaFlow (Wan2.2)}
    \end{subfigure}

    \caption{Failure-mode breakdown for the \emph{Brush} task.}
    \label{fig:brush_failure_modes}
\end{figure}

\begin{figure}[t]
    \centering

    \begin{subfigure}[t]{0.48\textwidth}
        \centering
        \includegraphics[width=\textwidth]{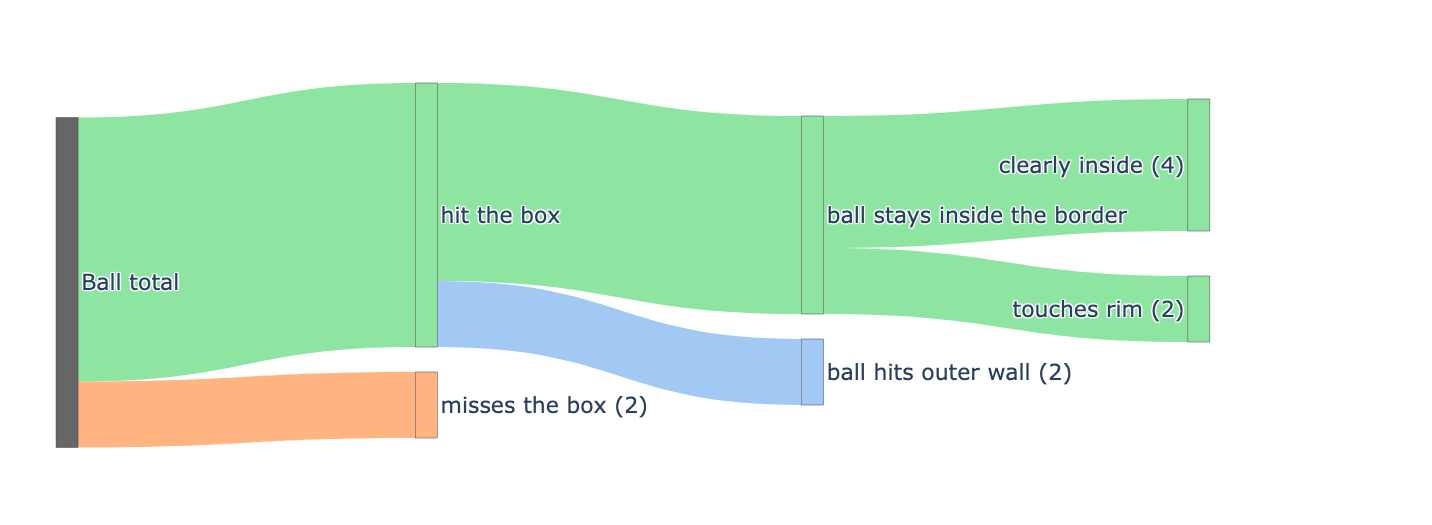}
        \caption{TraceGen (Warmed up with robot demo)}
    \end{subfigure}%
    \par\smallskip
    \begin{subfigure}[t]{0.48\textwidth}
        \centering
        \includegraphics[width=\textwidth]{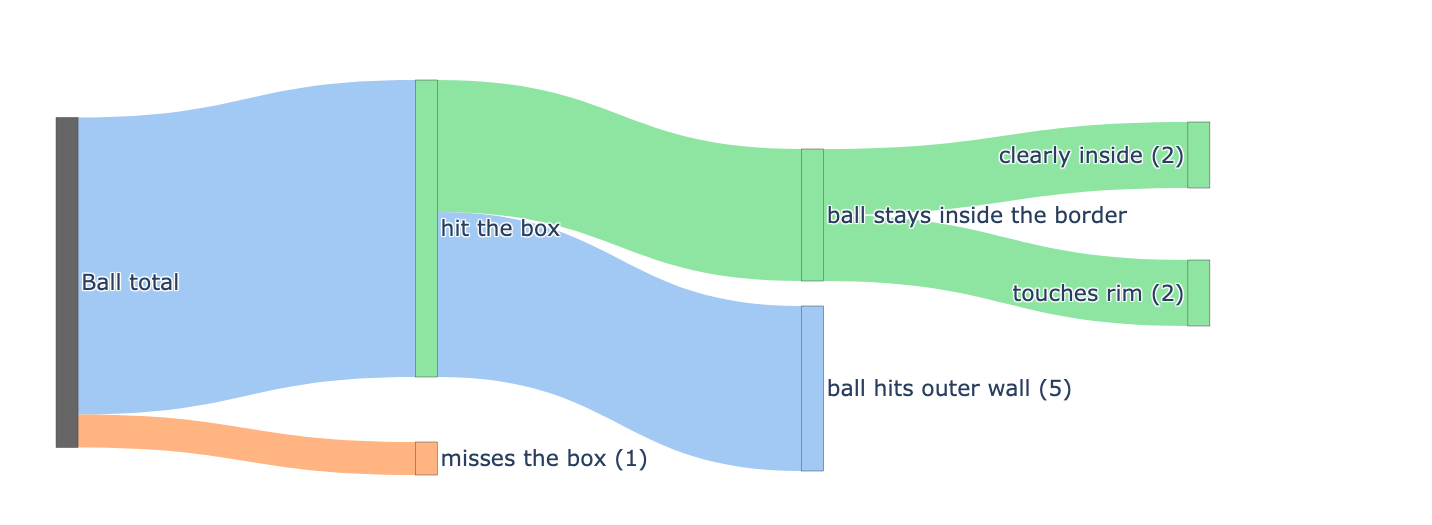}
        \caption{TraceGen (Warmed up with human demo)}
    \end{subfigure}

    \par\smallskip
    \begin{subfigure}[t]{0.48\textwidth}
        \centering
        \includegraphics[width=\textwidth]{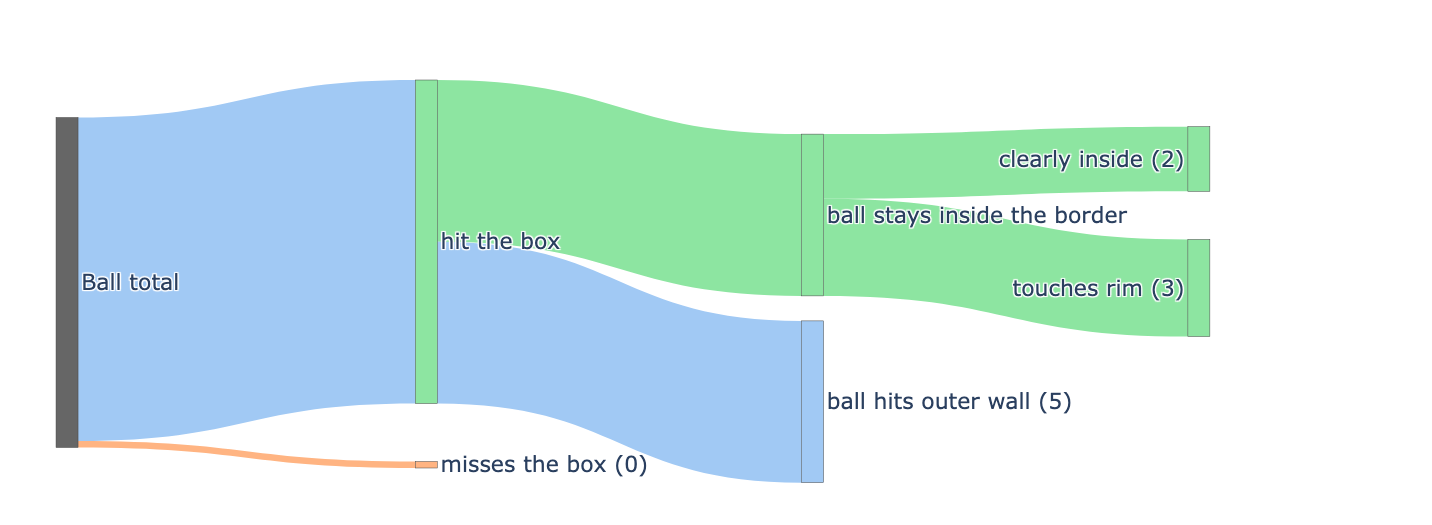}
        \caption{NovaFlow (Veo3.1)}
    \end{subfigure}%
    \par\smallskip
    \begin{subfigure}[t]{0.48\textwidth}
        \centering
        \includegraphics[width=\textwidth]{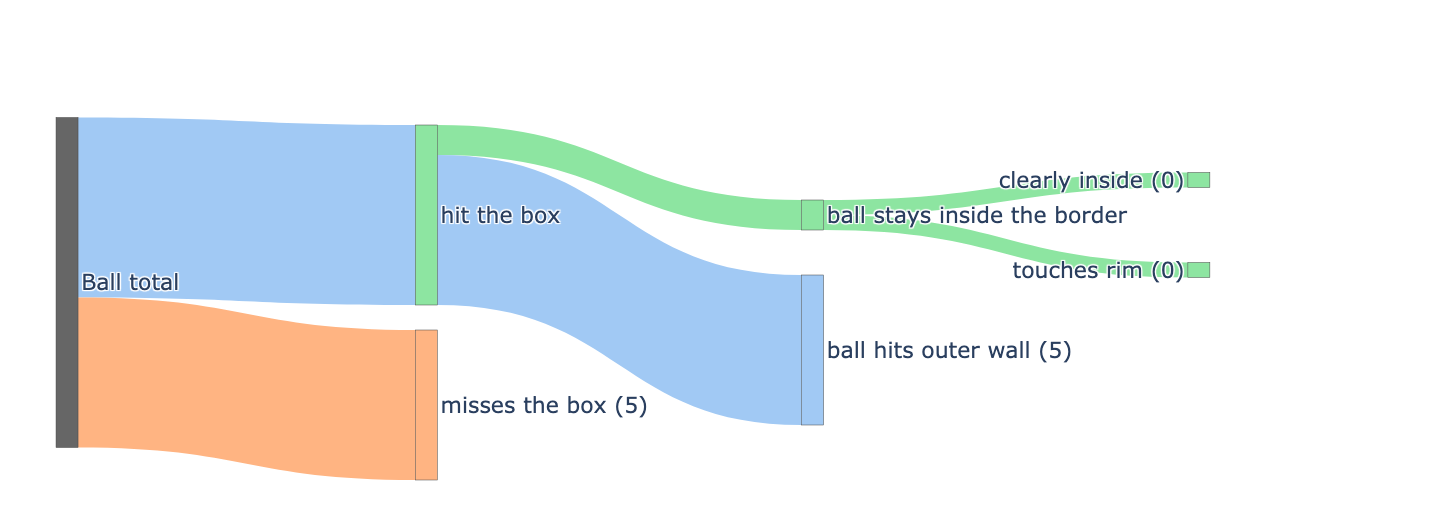}
        \caption{NovaFlow (Wan2.2)}
    \end{subfigure}

    \caption{Failure-mode breakdown for the \emph{Ball} task.}
    \label{fig:ball_failure_modes}
\end{figure}

\paragraph{NovaFlow}
The original NovaFlow pipeline includes a grasp-proposal module and a trajectory-planning module. Since our tasks do not involve grasping, we remove the grasp-proposal stage and initialize the robot in a state where it is already holding the object, ensuring a fair comparison.

In addition, the official NovaFlow implementation uses MegaSaM~\cite{li2025megasam} with TAPIP3D for camera pose estimation and depth prediction. For fair comparison, we replace the MegaSaM component with the same fine-tuned VGGT~\cite{vggt} depth and pose predictor from SpatialTrackerV2~\cite{xiao2025spatialtrackerv2} that we use in TraceForge. This VGGT-based predictor achieves similar accuracy while providing substantially faster inference by avoiding expensive 3D optimization. Also, for each NovaFlow, we use the following prompt:

\begin{itemize}
    \item Veo3.1 (Clothes): The robot smoothly picks up the pants leg and folds the garment in half.
    \item Veo3.1 (Brush): The robot arm moves the broom toward the yellow trash, sweeps it forward, and guides it into the dustpan.
    \item Veo3.1 (Block): In the picture you see robot, blue cube, red paper, and purple paper. The blue cube is right underneath the robot arm. These are the only things that you need to pay attention. The robot arm grabs the blue cube and put it on the top of the purple paper.
    \item Veo3.1 (Ball): The robot arm in the image moves to grab the tennis ball and put it into the box in the image.
    \item Wan2.2 (Clothes): The robot's end effector grasps the black pants, positions them flat, then folds them in half to create a compact folded shape.
    \item Wan2.2 (Brush): The robot's end effector grips the broom handle and sweeps the yellow trash into the dustpan with deliberate strokes..
    \item Wan2.2 (Block): The robot's end effector grasps the LEGO block, lifts it upward, moves it above the purple notebook, then lowers it onto the notebook center.
    \item Wan2.2 (Ball): The robot's end effector grasps the tennis ball, lifts it upward, then moves it horizontally toward the box and lowers it inside.
\end{itemize}

\paragraph{AVDC}
For AVDC, we follow the exact evaluation setup used in the NovaFlow paper. Specifically, we isolate the video-generation component from AVDC and combine it with the same 3D point-tracking and depth-estimation modules used in both TraceForge and NovaFlow, ensuring consistency across all baselines.

\paragraph{Inference latency measurement.}
All baseline runtimes were measured on an NVIDIA RTX A5000 except Wan~2.2 and Veo~3.1. 
Wan~2.2 could not fit on a single A5000, so we enabled inference using multiple GPUs---an unavoidable choice that in fact favors the baseline. 
Veo~3.1 is closed-source, and its latency is reported based on average API response time, which again places the baseline at an advantage.

\section{Failure modes analysis}

To better understand the behavior of each method beyond success rates, we provide a detailed failure‐mode analysis across all four tasks and four model configurations: (i) \textbf{TraceGen} with robot-domain warmup, (ii) \textbf{TraceGen} with human video warmup, (iii) \textbf{NovaFlow} (Veo3.1), and (iv) \textbf{NovaFlow} (Wan2.2).
For each combination of task and method, we collect every executed trial and categorize the outcome into fine-grained success and failure types based on object interaction quality and task completion criteria.

We visualize these distributions in Sankey diagrams, which reveal how trials
progress from the total pool of attempts (left) into distinct outcome modes
(right). This representation highlights the long-tail structure of
failure patterns---showing whether errors arise early (e.g., incorrect approach) or late in the trajectory (e.g., partial completion, drift during final alignment).

\subsection{Warmup data}\label{warmup_data}

In this section, we visualize all warmup demonstrations used in our experiments. As described in the main text, TraceGen is adapted to each
task using a lightweight warmup stage, which serves to translate the
embodiment-agnostic 3D traces into the action space of the target robot or tasks.

\begin{figure}[t!]
    \centering
    \includegraphics[width=\linewidth]{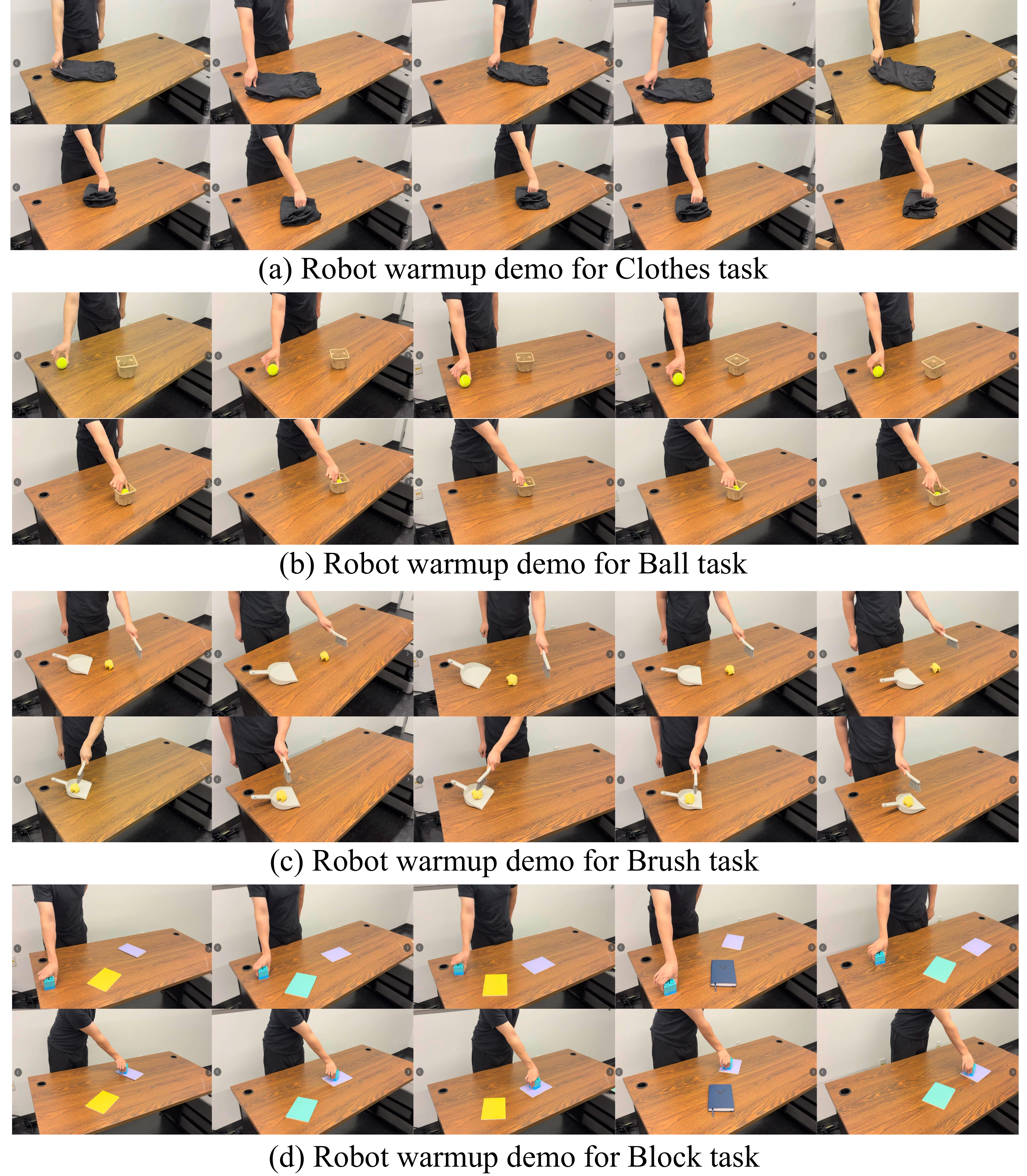}
    \caption{Human warmup demonstrations for all four tasks, showing first (top) and final (bottom) frames of each handheld video.}
    \label{fig:warmup_human}
\end{figure}
\begin{figure}[t!]
    \centering
    \includegraphics[width=\linewidth]{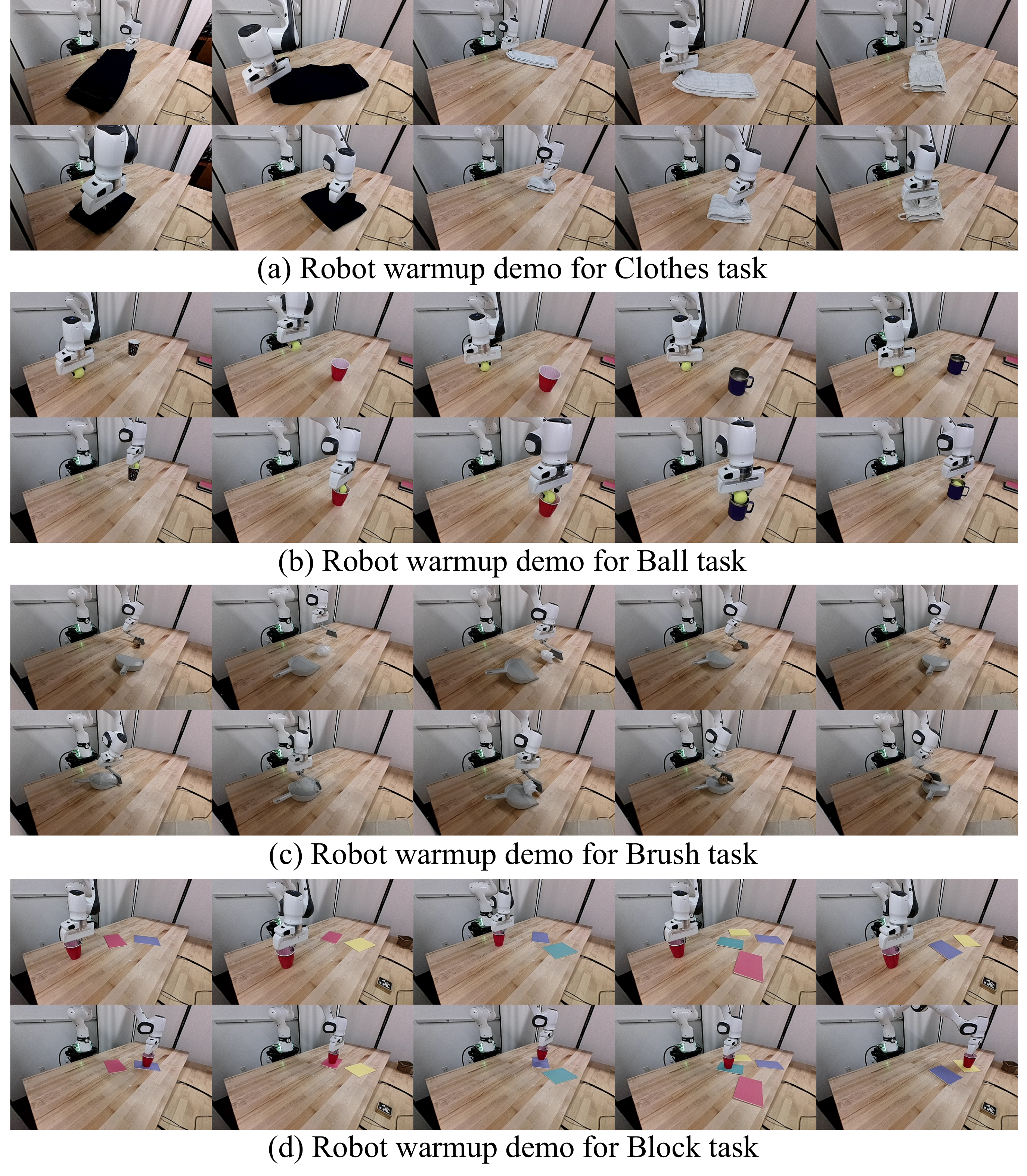}
    \caption{Robot warmup demonstrations for all four tasks. Top row shows the first frame of each demo; bottom row shows the final frame.}
    \label{fig:warmup_robot}
\end{figure}
We consider two warmup regimes:
\begin{itemize}[leftmargin=*, itemsep=2pt]
    \item \textbf{Robot$\rightarrow$Robot (same-embodiment warmup).}
    Five in-domain robot demonstrations are provided for each task.
    These clips differ from the evaluation setting in object layout and initial robot pose, ensuring that warmup does not simply memorize target configurations.
    \item \textbf{Human$\rightarrow$Robot (cross-embodiment warmup).}
    Five uncalibrated human videos (3–4 seconds each) are captured per task using a handheld phone. These clips differ substantially from the robot setting in background, lighting, embodiment, and object placement.
\end{itemize}

Each figure below shows all warmup demonstrations for the four tasks
(\emph{Clothes}, \emph{Block}, \emph{Brush}, \emph{Ball}). For each demo,
the \textbf{top row displays the first frame} and the \textbf{bottom row
displays the final frame}.

\subsection{Long horizon experiments}

To assess whether TraceGen’s predicted trace can be composed into longer multi-step behaviors, we evaluate the model on a long-horizon \emph{Sorting} task. The goal is to separate blocks from white trash items: each block must be placed in a designated green region, and each trash item must be placed in the red region. We collect five human teleoperation demonstrations of the full sorting process and segment them into four
primitive subtasks, which serve as warmup data for TraceGen.

The sorting procedure consists of the following four consecutive subtasks:
\begin{enumerate}
    \item Place the left trash on the red paper.
    \item Place the pink LEGO block on the green paper.
    \item Place the blue LEGO block on the green paper.
    \item Place the right trash on the red paper.
\end{enumerate}
Completing all four subtasks in sequence constitutes a successful sorting
episode.

\paragraph{Use of scripted grasping.}
Because TraceGen models only the 3D trace component of manipulation and
does not include an external grasping module, we assume access to a
pre-defined scripted policy for picking up each object. This policy moves the robot to a preset grasping pose, enabling the placing skill generated by TraceGen to begin from a consistent home configuration.

\paragraph{Results.}
We compare TraceGen initialized with pretraining from the TraceForge-123k dataset (``Warmed up from
TraceGen'') against a \emph{From Scratch} model trained only on the four warmup
segments. Table~\ref{tab:sorting_long_horizon} reports per-step success rates
across 10 rollouts. While the pretrained model occasionally fails the first
trash placement, it maintains high performance on all subsequent subtasks.
In contrast, the scratch model exhibits compounding errors over time, with
success rates degrading markedly in the later steps.

\begin{table}[h]
\centering
\footnotesize
\caption{Long-horizon Sorting task: per-subtask success rates (left to right
indicates temporal order).}
\label{tab:sorting_long_horizon}
\resizebox{\columnwidth}{!}{%
\begin{tabular}{lcccc}
\toprule
\textbf{Model} 
& \textbf{Left Trash} $\rightarrow$
& \textbf{Pink Block} $\rightarrow$
& \textbf{Blue Block} $\rightarrow$
& \textbf{Right Trash} \\
\midrule
Warmed up from TraceGen & 0.8 & 0.8 & 0.8 & 0.8 \\
From Scratch             & 1.0 & 0.8 & 0.5 & 0.4 \\
\bottomrule
\end{tabular}
}
\end{table}

\begin{figure}[t!]
    \centering
    \includegraphics[width=\linewidth]{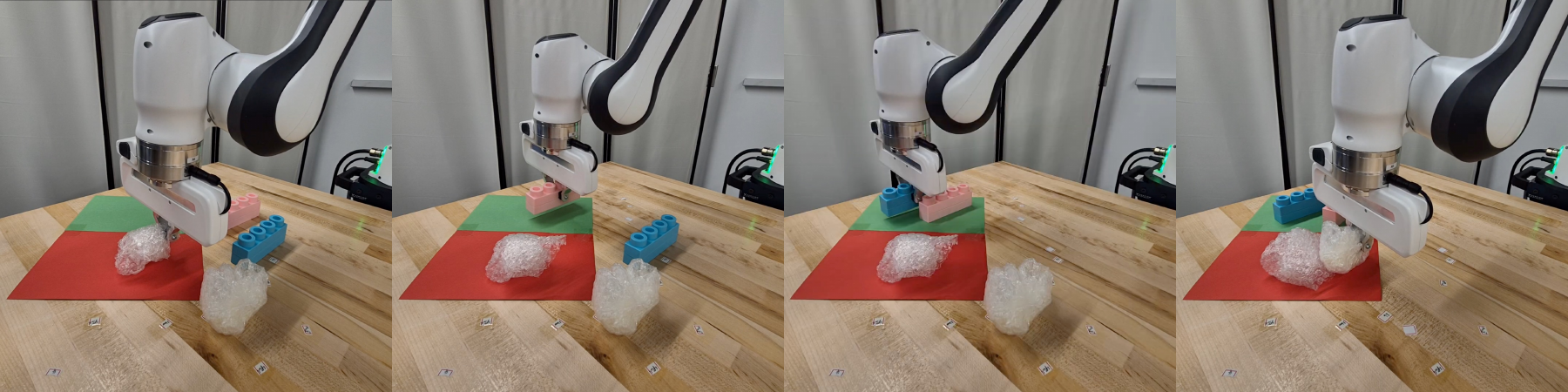}
    \caption{Visualization of the long-horizon Sorting task, showing the four sequential placement subtasks from left to right.}
    \label{fig:longhorizon}
\end{figure}

Overall, these long-horizon results show that TraceGen’s pretrained
motion priors enable stable composition of primitive placing behaviors,
mitigating error accumulation across sequential subtasks. Although the
model was not explicitly optimized for extended planning, its compact 3D
trace representation supports reliable stitching of skills over longer
task horizons.

\end{document}